\documentclass{article}
\usepackage{colm2024_conference}
\usepackage[most]{tcolorbox}
\usepackage[font=small,labelfont=bf]{caption}
\usepackage{amsmath,amssymb,booktabs,tabularx,array}
\usepackage{graphicx,xcolor,tikz}
\usepackage{colortbl}
\usepackage{algorithm}
\usepackage{algpseudocode}
\usepackage{placeins}
\usetikzlibrary{arrows.meta,positioning,calc,fit}
\usepackage{hyperref}
\usepackage{url}

\title{\raisebox{-0.18em}[0pt][0pt]{\includegraphics[height=1.45em]{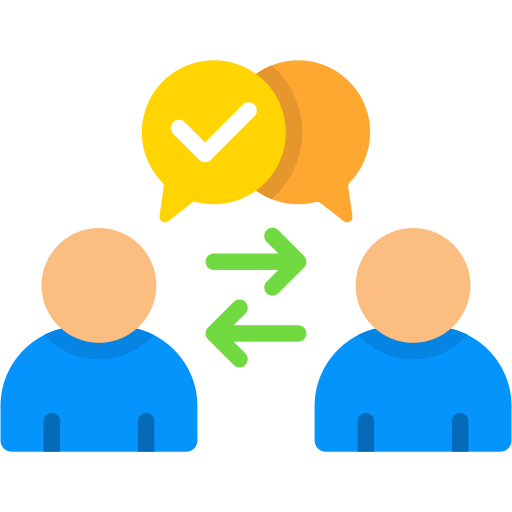}}\hspace{0.35em}What Makes a Social Agent's Interactions Effective?\\[0.35em]
Learning from Verifiable Outcomes in Paired Worlds}
\hypersetup{pdftitle={What Makes a Social Agent's Interactions Effective? Learning from Verifiable Outcomes in Paired Worlds},pdfauthor={Caixin Kang, Ke Wang, Yuchuan Wu, Xiaoqiang Wang, Ling Li, Bingxue Qiu, Yongbin Li, Yoichi Sato}}
\author{%
{\fontsize{11}{14}\selectfont
Caixin Kang\textsuperscript{1,2},
Ke Wang\textsuperscript{1},
Yuchuan Wu\textsuperscript{1},
Xiaoqiang Wang\textsuperscript{1},\\
Ling Li\textsuperscript{1},
Bingxue Qiu\textsuperscript{1},
Yongbin Li\textsuperscript{1}\thanks{Corresponding author.},
Yoichi Sato\textsuperscript{2}}\\[0.5em]
{\large \textsuperscript{1}Alibaba Token Hub, Alibaba Group\\
\textsuperscript{2}The University of Tokyo}}

\newcommand{\bench}{HICO-Bench}
\newcommand{\method}{PaCPO}
\newcolumntype{Y}{>{\raggedright\arraybackslash}X}

\definecolor{qwenink}{HTML}{252A4A}
\definecolor{qwenmuted}{HTML}{69708F}
\definecolor{qwengrid}{HTML}{DDE2F4}
\definecolor{qwenlavender}{HTML}{C9CEFF}
\definecolor{qwencyan}{HTML}{55B8F6}
\definecolor{qwenblue}{HTML}{6477EE}
\definecolor{qwenviolet}{HTML}{8B72E8}
\definecolor{qwenpurple}{HTML}{5B50D6}
\definecolor{qwendeep}{HTML}{37308F}
\definecolor{tablegray}{HTML}{EEEEEE}
\definecolor{citesky}{HTML}{299DD8}
\hypersetup{colorlinks=true,linkcolor=qwendeep,citecolor=qwendeep,urlcolor=qwenblue}
\newcommand{\tabref}[1]{{\scriptsize\citep{#1}}}

\colorlet{mypurple}{qwenpurple}
\colorlet{deeppurple}{qwendeep}
\fancypagestyle{firstpage}{%
  \fancyhf{}%
  \fancyhead[L]{\raisebox{-3pt}[20pt][0pt]{\includegraphics[height=24pt]{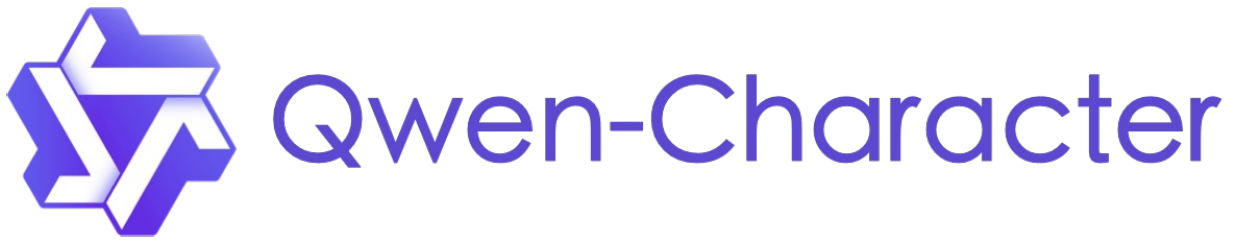}}}%
  \fancyfoot[C]{\thepage}%
}
\begin{document}
\maketitle

\begin{abstract}
Effective social interaction requires more than fluent responses: agents
must reach plans that fit the participants' needs and circumstances.
Existing evaluations of social agents mainly rely on LLM judges that rate
the conversation, without checking the resulting plan. We introduce
\bench{}, a benchmark of 180 everyday coordination scenarios, such as
resource allocation, scheduling, and authorization, in which two agents
with private information converse and then independently record the plan
they agreed on. A program checks \emph{Feasibility} (the two records match
and satisfy both participants' private conditions), and an LLM judge
checks \emph{Integrity} (the conversation supports the recorded plan);
passing both is a \emph{Verified Success}. Each scenario has two versions
that share the same public context but differ in one private condition,
so the correct plan differs between them. Across thirteen LLMs, 36\% of
conversations rated highly for goal completion fail Verified Success. To
learn from outcomes that arrive only at the end of a conversation, we
propose Paired-Credit Policy Optimization (\method{}). After each turn,
\method{} asks the agent for a tentative plan, checks it in both versions
of the scenario, and rewards the turns after which the tentative plans
fit their respective versions. \method{} raises Qwen3-8B's Verified
Success from 20.4\% to 28.3\%.
\end{abstract}

\section{Introduction}
\label{sec:introduction}

LLM agents increasingly help people coordinate everyday
matters. Consider Maya asking her colleague Leo, \emph{``Could you give me
a hand with tomorrow's client pitch?''} (Figure~\ref{fig:overview}).
Whether Leo should review her slides or present in her place depends on
facts that only Maya knows: whether she can attend, and what help she
wants. A reply such as \emph{``Don't worry, I'll take over''} sounds
supportive, yet commits to a plan that fails if Maya can present herself.
Splitting a bill, scheduling a meeting, and approving a request share
this structure: an acceptable outcome depends on \emph{private
conditions}, such as budget, availability, or authority, revealed
through dialogue.

Existing evaluations of LLM social agents focus primarily on the
conversation. SOTOPIA \citep{zhou2023sotopia} uses an LLM judge to rate
agents on goal completion, believability, and related dimensions, and
subsequent work trains agents to improve these ratings
\citep{wang2024sotopiapi,yu2025sotopiarl}. Such ratings do not verify
whether the agreed plan satisfies the participants' conditions, and
model-based judges can overestimate agents' social abilities
\citep{wang2024sotopiapi}. Negotiation and tool-use benchmarks do verify
outcomes automatically \citep{lewis2017deal,yao2024taubench}, but not in
everyday coordination, where the decisive facts must be elicited through
conversation.

We introduce \bench{} (Hidden Information and Checkable Outcomes), a
benchmark that evaluates social interaction by its outcome. After each
conversation, both agents independently record
the agreed plan in a structured form. A program checks
\emph{Feasibility}: whether the two records match and satisfy all private
conditions. An LLM judge checks \emph{Integrity}: whether the
conversation supports the recorded plan, for example without invented
authority or a record that contradicts what was said. An interaction is
a \emph{Verified Success} only if it passes both checks. Each scenario
has two versions, or \emph{worlds}, that differ in a single private
condition (e.g., whether Maya can attend) and admit disjoint sets of
correct plans; an agent that responds identically in both worlds
therefore fails at least one. Across thirteen LLMs, 36\% of conversations
with Goal $\geq 8/10$ fail Verified Success.

Verified outcomes can also serve as a training signal, but they arrive
only after the final turn, leaving the contribution of individual turns
unclear. We propose \emph{Paired-Credit Policy Optimization} (\method{}),
which uses the paired worlds for credit assignment. After each turn, we
probe the agent, outside the conversation, for a tentative plan it would
record at that point, and check this plan in both worlds. Before
Leo learns whether Maya can attend, his tentative plan is the same in
both worlds and can fit at most one; once she answers, the plan formed
in each world should fit that world but not its counterpart. \method{}
rewards the turns at which this occurs, redistributing the terminal
reward without changing each conversation's total return. On Qwen3-8B,
\method{} improves Verified Success from 20.4\% to 28.3\%, compared with
25.0\% for PPO and 26.3\% for GRPO trained on the same number of
dialogues. Our contributions are as follows:

\begin{figure}[t]
\centering
\includegraphics[width=\linewidth]{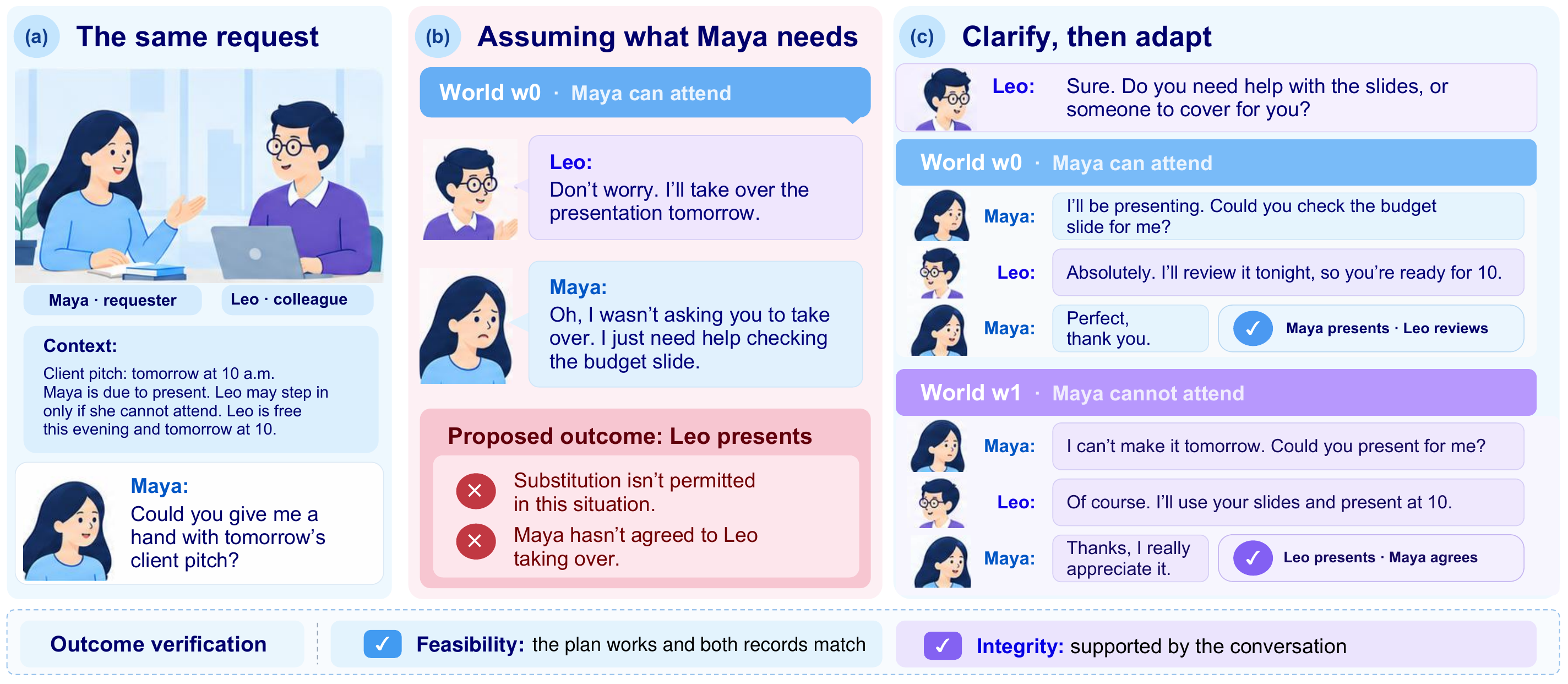}
\caption{\textbf{Reassuring words need not lead to appropriate help.} (b) Without clarifying, Leo offers to take over, although Maya can present and only wants a slide review. (c) After clarifying, the appropriate plan depends on whether Maya can attend, which is initially private to her. Feasibility checks that the plan works and both records match; Integrity checks that dialogue supports it.}
\label{fig:overview}
\end{figure}

\begin{itemize}
  \item \textbf{\bench{}}, a benchmark of 180 paired scenarios across six
  coordination families, evaluated by rule-based Feasibility and
  judge-based Integrity alongside SOTOPIA ratings.
  \item \textbf{\method{}}, a credit-assignment method that compares
  tentative plans across paired worlds once the private condition that
  differs between them has been disclosed.
  \item \textbf{An evaluation of thirteen LLMs and five RL baselines},
  showing that highly rated conversations can yield failed plans and
  that outcome training improves Verified Success.
\end{itemize}

\section{Related Work}
\label{sec:related}

\paragraph{Agent reasoning and interaction.}
Within broader research on agent reasoning and interaction
\citep{yao2023react,liu2025advances,wang2025oscar,wang2026system},
Social IQa and FANToM test social commonsense and who knows what in a
conversation \citep{sap2019socialiqa,kim2023fantom}.
SOTOPIA evaluates agents in role-played social scenarios, with an LLM
judge rating goal completion, believability, and other dimensions
\citep{zhou2023sotopia}; SOTOPIA-ToM extends this setting to information
seeking and privacy \citep{ys2026sotopiatom}.
Multimodal work also studies truthfulness and deception
\citep{kang2025truthfulness,kang2025can}, personality judgments
\citep{kang2026perception}, and speaker alignment
\citep{ouyang2026multi}, alongside interactive digital humans
\citep{cai2025interactive} and controllable multi-person video generation
\citep{ouyang2026socialdirector}.
\bench{} focuses on the plan reached through dialogue: whether it fits
both participants' private conditions and is supported by what they said.
We retain SOTOPIA's ratings as complementary measures of conversation quality.

\paragraph{Task-oriented dialogue and negotiation.}
Several interactive benchmarks verify outcomes with rules. In Deal or No
Deal and CaSiNo, agents divide items and the final split is scored
against each agent's private values \citep{lewis2017deal,chawla2021casino};
$\tau$-bench and $\tau^2$-bench check whether a customer-service agent
leaves a database in the correct state \citep{yao2024taubench,barres2025tau2};
Social Gym uses games whose winner is decided by rules
\citep{he2026socialgym}. These settings target fixed,
domain-specific goals. \bench{} extends outcome verification to everyday
coordination and additionally checks that the conversation supports the
plan.

\paragraph{Clarification and information seeking.}
Prior work trains or evaluates models on asking clarifying questions,
for example by rewarding questions that reveal user preferences
\citep{andukuri2024stargate} or by scoring questions at checkpoints
\citep{zhao2026askbench,zhang2024clamber}. Related work on proactive
assistance studies when to offer help from ongoing visual observations
\citep{gong2026vinci2,gong2026grove}. We do not reward questions
directly: in \bench{}, the decisive fact may be volunteered, already
known, or elicited, and only the correctness of the resulting plan is
rewarded.

\paragraph{Credit assignment in reinforcement learning.}
When feedback arrives only at the end of an episode, a learner must
estimate which earlier steps deserve credit. Process supervision labels each reasoning
step \citep{lightman2024verify}; VinePPO estimates a step's value by
sampling continuations from it \citep{kazemnejad2025vineppo}; GiGPO and
GraphGPO compare steps that share the same state across rollouts
\citep{feng2025gigpo,cheng2026graphgpo}. For social agents, SOTOPIA-RL
and SAVOIR distribute LLM-judge ratings over the utterances of a
conversation \citep{yu2025sotopiarl,feng2026savoir}. In free-form
dialogue, however, identical states rarely recur across rollouts, and
LLM ratings are noisy. \method{} instead compares tentative plans across
the two worlds of a scenario and checks them with rules, so that each
comparison is verifiable.

\section{HICO-Bench}
\label{sec:task}

\subsection{Interaction Protocol}

Each scenario involves two roles, such as Maya and Leo in
Figure~\ref{fig:overview}. Both roles see the same \emph{public context}
$p$: the client pitch is tomorrow at 10 a.m., Maya is due to present,
and Leo may present only if Maya cannot attend. Each role $i$ also
receives a private \emph{role card} $v_i$ that the other role does not
see; Maya's card states whether she can attend and what help she wants.
The roles then alternate free-form turns, producing a
history $h$. Each role observes only $p$, its own card, and $h$; it never
sees the partner's card or the set of acceptable plans.

When the conversation ends, each role independently fills in a
structured \emph{record} $c_i$ describing the plan it believes was
agreed, following a public template $\mathcal{C}$ (e.g., who presents
and what Leo will do). Independent records test whether the roles reached a
shared understanding \citep{clark1991grounding}. The \emph{verifier
state} $z$ stores the private conditions of both roles. A scenario is thus
$x=(p,v_1,v_2,z,\mathcal{C})$.

\subsection{Paired Scenarios}

Each scenario has two worlds, $w_0$ and $w_1$, that share the public
context and differ in exactly one private condition, chosen so that no
plan is correct in both.
In our example, Maya can attend in $w_0$, so the correct plan is that she
presents and Leo reviews her slide; in $w_1$ she cannot attend, so the
correct plan is that Leo presents. Following contrast sets
\citep{ribeiro2020checklist,gardner2020contrastsets}, an agent that makes
the same offer in both worlds must fail one of them.

Table~\ref{tab:families} lists the six coordination families. We adapt
social situations from SOTOPIA \citep{zhou2023sotopia}, write role cards
for both worlds, and encode the acceptance rules as a program, which is
tested on correct and incorrect plans. The 180 scenarios are split into
96 training, 24 development, and 60 test scenarios, with both worlds in
the same split. Evaluating each role in each world separately yields
$60\times2\times2=240$ test tasks (96 development tasks).

\begin{table}[t]
\centering\small\setlength{\tabcolsep}{6pt}\renewcommand{\arraystretch}{1.14}
\caption{\textbf{Interaction families in HICO-Bench.} Each family has 16 training, 4 development, and 10 test scenarios, each with two worlds.}
\label{tab:families}
\begin{tabularx}{\linewidth}{@{}>{\raggedright\arraybackslash}p{.23\linewidth}>{\raggedright\arraybackslash}p{.29\linewidth}Y@{}}
\toprule
\textbf{Interaction family} & \textbf{Example} & \textbf{Private conditions and verification}\\
\midrule
\textbf{Resource allocation} & Neighbors divide a \$240 fence-repair bill. & Shares respect both participants' bounds and sum to the fixed total.\\
\addlinespace[4pt]
\textbf{Authorization} & A consultant handles a request for one extra review hour. & Approval stays within delegated authority; otherwise, the request is referred.\\
\addlinespace[4pt]
\textbf{Bounded quantities} & A buyer and seller agree on the number of apple crates. & The agreed integer quantity lies within both participants' acceptable intervals.\\
\addlinespace[4pt]
\textbf{Joint scheduling} & Studio partners arrange a 20-minute celebration. & The shared slot fits both availability windows and has the required duration.\\
\addlinespace[4pt]
\textbf{Shared-resource scheduling} & Two photographers reserve a tripod for ten minutes each. & Bookings fit both use windows and required durations, without overlapping.\\
\addlinespace[4pt]
\textbf{Set-based consent} & Two people agree on belief-related discussion topics. & Selected topics match the required set and are all permitted.\\
\bottomrule
\end{tabularx}
\end{table}

\subsection{Outcome Evaluation}
\label{sec:verification}

\paragraph{Feasibility: does the plan work?}
Each record specifies an outcome \emph{kind} and its \emph{terms}. The
kind is \emph{agreement} (a plan is settled), \emph{decline} (no plan
satisfies both roles), or \emph{defer} (the request requires approval
from someone outside the conversation). In $w_0$ of our example, a
correct record is
{\small\texttt{\{kind: agreement, terms: \{presenter: Maya, leo\_task: review\_budget\_slide\}\}}}.
A program checks that (1) the two records are identical up to the
ordering of fields and items; (2) the kind is correct for the world; and
(3) the terms satisfy the family's rules (Table~\ref{tab:families}), e.g.,
in resource allocation the shares sum to the \$240 bill and respect each
person's limit. Any plan satisfying the rules is accepted; in scheduling,
for instance, any mutually free slot passes. $F_z(c_1,c_2)=1$ if all
three tests pass and 0 otherwise.\par

\paragraph{Integrity: does the conversation support the plan?}
A feasible record may still be unsupported by the conversation: both
records may state that Maya presents while Leo promised to take over, or
Leo may approve extra work by claiming authority he lacks. An LLM judge
reads the full conversation and both records and flags four violations,
each with quoted evidence: (i) an uncorrected factual error, such as
invented authority or resources; (ii) repeated exchanges without
progress; (iii) a record that contradicts the spoken agreement;
and (iv) output-format text leaking into the dialogue.
$I_z(h,c_1,c_2)=1$ if no violation is found. Corrected proposals and
truthful referrals are allowed.

\paragraph{Verified Success.}
Our primary metric is the fraction of tasks that pass both checks,
\begin{equation}
S_z(h,c_1,c_2)=F_z(c_1,c_2)\,I_z(h,c_1,c_2).
\label{eq:joint}
\end{equation}
We also report each check separately, as they capture different
failure modes.

\paragraph{Conversation ratings.}
We additionally report SOTOPIA's seven LLM-judged ratings
\citep{zhou2023sotopia}, focusing on \emph{Goal} (0--10, attainment of
the role's social goal) and \emph{Believability} (0--10, naturalness and
role consistency); all seven appear in Appendix~\ref{app:metrics}. These
ratings are never used for training; they test whether high ratings
coincide with verified outcomes.

\section{Paired-Credit Policy Optimization}
\label{sec:method}

\begin{figure}[t]
\centering
\includegraphics[width=\linewidth]{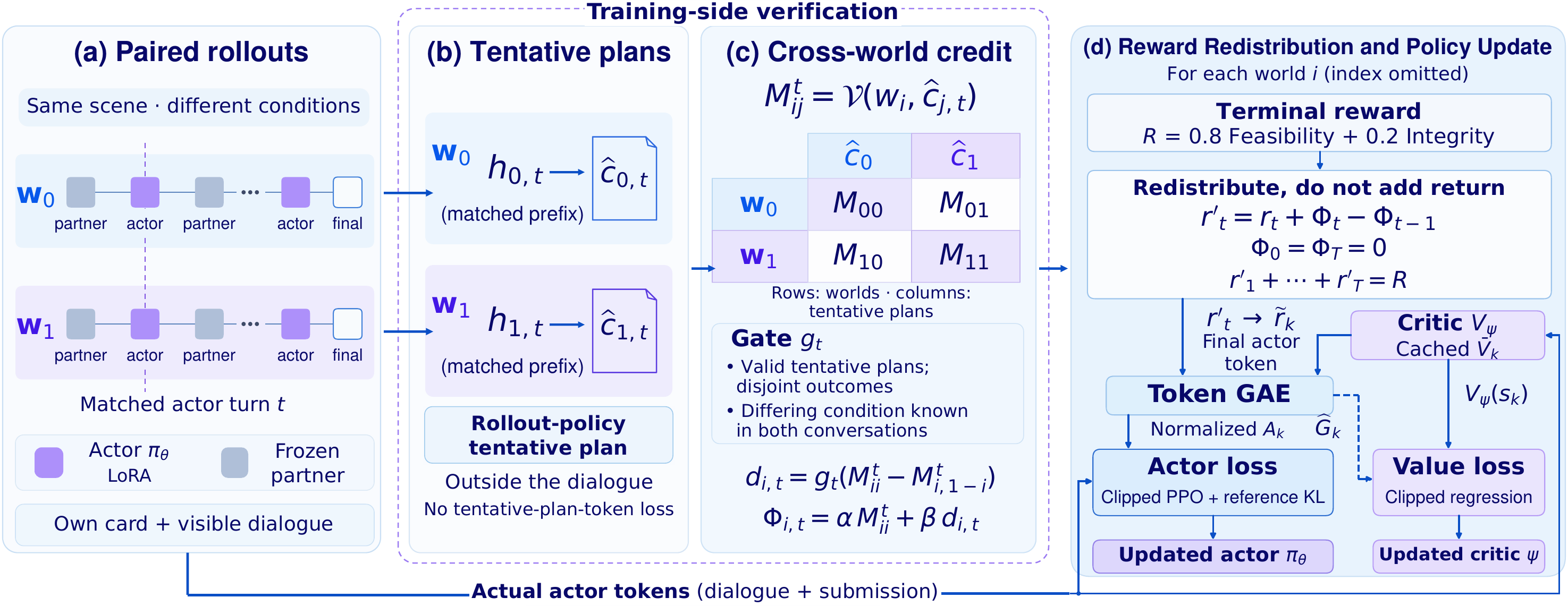}
\caption{\textbf{PaCPO overview.} Rollouts in the two worlds yield a tentative plan after each turn. Checking each plan in both worlds determines turn-level credit, which redistributes the terminal reward before the PPO update. Indices $i,j$ denote worlds, $t$ turns, and $k$ tokens; panel (d) omits $i$. $R$ is the terminal reward, $r_t$ and $r'_t$ are the original and redistributed rewards, and $\Phi_t$ is the potential. The dashed branch carries GAE value targets.}
\label{fig:pacpo-training}
\end{figure}

\method{} trains the agent with a terminal reward but redistributes it
across turns. After each turn, we ask which plan the agent would record
if the conversation ended there, and compare these tentative plans across
the two worlds of a scenario. Before Maya reveals whether she can attend,
Leo's tentative plan is the same in both worlds and can fit at most one;
once she does, the plans formed in the two worlds should diverge so that
each fits its own world. \method{} rewards the turns at which this
divergence appears. Figure~\ref{fig:pacpo-training} illustrates the
pipeline; Algorithm~\ref{alg:pacpo} (Appendix~\ref{app:implementation})
details the update.

\subsection{Terminal Reward}

All RL methods in our experiments optimize the same terminal reward
\begin{equation}
R(h,c_1,c_2,z)=0.8F_z(c_1,c_2)+0.2I_z(h,c_1,c_2),
\label{eq:reward}
\end{equation}
which gives partial credit when only one check passes. PPO and GRPO
receive $R$ at the final turn; \method{} redistributes it across turns.

\subsection{Tentative Plans and Cross-World Checks}
\label{sec:probes}

Let $i\in\{0,1\}$ index the worlds. For each scenario, we roll out the
same tested role in both worlds against a fixed partner; the two
conversations may differ in wording and length. Let $v_i$ be the tested
role's card in world $w_i$, $h_{i,t}$ the conversation up to its $t$-th
turn, and $T_i$ the turn at which it submits its record. After each turn
$t<T_i$, we probe the current policy for a \emph{tentative plan}: the plan
it would record if the conversation ended there, given only what the
role can see:
\begin{equation}
\widehat c_{i,t}\sim\pi_{\theta_{\mathrm{old}}}
(\cdot\mid p,v_i,h_{i,t},\mathcal{C}).
\label{eq:probe}
\end{equation}
The tentative plan is not added to the conversation.

Let $\mathcal{V}(w,c)\in\{0,1\}$ indicate whether plan $c$ satisfies the
Feasibility rules of world $w$, with $\mathcal V(w,\bot)=0$ for a missing
or malformed plan. At turn $t$, \method{} checks both tentative plans in
both worlds:
\begin{equation}
M_{ij}^{t}=\mathcal{V}(w_i,\widehat c_{j,t}),
\qquad i,j\in\{0,1\}.
\label{eq:crossverify}
\end{equation}
The diagonal entries indicate whether each world's plan fits its own
world; the off-diagonal entries indicate whether the plan from the other
world would also fit.

\subsection{Paired Credit}

The comparison is informative only after the private condition that
differs between the worlds has been disclosed. We therefore compare plans
at the same turn index and open a gate, $g_t=1$, only if both tentative
plans are valid, the two worlds admit no common plan, and the differing
condition is available in both conversations by turn $t$: it is either
on the tested role's own card or has been disclosed by the partner, as
marked by a frozen LLM annotator during training. For world $i$, the
paired credit is
\begin{equation}
d_{i,t}=\begin{cases}
M_{ii}^{t}-M_{i,1-i}^{t}, & g_t=1,\\
0, & g_t=0.
\end{cases}
\label{eq:pairedcredit}
\end{equation}
It is positive when the plan formed in $w_i$ fits $w_i$ but the plan from
the other world does not. In our example, once Maya has answered,
proposing that Maya presents in $w_0$ and that Leo presents in $w_1$
gives $M^t=\bigl[\begin{smallmatrix}1&0\\0&1\end{smallmatrix}\bigr]$ and
hence $d_{0,t}=d_{1,t}=1$; identical plans in both worlds give zero
credit. The potential at turn $t$ combines within-world feasibility and
paired credit:
\begin{equation}
\Phi_{i,t}=\alpha M_{ii}^{t}+\beta d_{i,t}.
\label{eq:potential}
\end{equation}

\subsection{Reward Redistribution}

Let $r_{i,T_i}=R_i$ and $r_{i,t}=0$ for $t<T_i$, and set
$\Phi_{i,0}=\Phi_{i,T_i}=0$. Following potential-based shaping and return
redistribution \citep{ng1999shaping,arjona2019rudder}, we define
\begin{equation}
r'_{i,t}=r_{i,t}+\Phi_{i,t}-\Phi_{i,t-1}.
\label{eq:redistribute}
\end{equation}
With $\gamma=1$, the shaping terms telescope, so the total reward of each
conversation is unchanged:
\begin{equation}
\sum_{t=1}^{T_i}r'_{i,t}=R_i+\Phi_{i,T_i}-\Phi_{i,0}=R_i.
\label{eq:conservation}
\end{equation}
A turn is thus rewarded when it raises the potential, for example by
eliciting the condition that lets the two worlds' plans diverge
correctly, and the final turn receives $R_i$ minus the credit already
assigned. We optimize the redistributed rewards with PPO
\citep{schulman2017ppo}, placing each $r'_{i,t}$ on the last token of
turn $t$ and using a learned value baseline with GAE
\citep{schulman2015gae}; Appendix~\ref{app:implementation} gives the
details.

\begin{table}[t]
\centering\small\setlength{\tabcolsep}{3pt}\renewcommand{\arraystretch}{1.08}
\caption{\textbf{Thirteen-model HICO-Bench benchmark.} Fixed Qwen3-8B partner, 240 tasks per model; sorted by increasing Verified Success (Ver. Succ.). Size denotes local model size; \texttt{api} denotes remote inference. Goal and Believability range from 0 to 10; full ratings are in Appendix~\ref{app:benchmark}.}\label{tab:benchmark}
\begin{tabularx}{\linewidth}{@{}lc*{5}{>{\raggedleft\arraybackslash}X}@{}}
\toprule
Model & Size & \multicolumn{3}{c}{Verifiable outcomes (\%)} & \multicolumn{2}{c}{SOTOPIA ratings}\\
\cmidrule(lr){3-5}\cmidrule(l){6-7}
 & & Feasibility & Integrity & \mbox{Ver. Succ.} & Goal & Believ.\\
\midrule
Qwen3-8B \tabref{yang2025qwen3} & 8B & 49.58 & 25.83 & 20.42 & 7.47 & 6.31\\
MiniMax M3 \tabref{minimax2026m3} & \texttt{api} & 58.33 & 62.50 & 50.42 & 9.18 & 8.68\\
GPT-4o \tabref{openai2024gpt4o} & \texttt{api} & 62.08 & 63.33 & 51.25 & 9.12 & 8.71\\
GPT-5.6 Luna \tabref{openai2026gpt56} & \texttt{api} & 73.75 & 56.67 & 52.08 & 9.61 & 8.63\\
Qwen3.8-27B \tabref{qwen2026qwen3827b} & 27B & 64.17 & 62.92 & 54.58 & 9.21 & 8.72\\
DeepSeek-V4-Pro \tabref{deepseek2026v4pro} & \texttt{api} & 68.33 & 66.25 & 59.17 & 9.44 & 8.88\\
Qwen3.8-Max \tabref{alibaba2026qwen38max} & \texttt{api} & 72.50 & 69.58 & 60.83 & 9.62 & 8.95\\
Gemma 4 31B IT \tabref{google2026gemma4} & 31B & 72.08 & 72.92 & 63.75 & 9.38 & 8.80\\
Kimi-K3 \tabref{moonshot2026kimik3} & \texttt{api} & 75.00 & 77.92 & 66.25 & 9.52 & 8.94\\
GLM-5.3 \tabref{zai2026glm53} & \texttt{api} & 73.75 & 75.83 & 67.08 & 9.65 & 8.96\\
GPT-5.6 Sol \tabref{openai2026gpt56} & \texttt{api} & 81.25 & 82.92 & 76.25 & 9.63 & 8.95\\
GPT-6 Astra \tabref{openai2026gpt6astra} & \texttt{api} & \textbf{87.50} & 86.25 & 80.83 & \textbf{9.75} & \textbf{9.01}\\
Gemini 3.7 Flash \tabref{google2026gemini37} & \texttt{api} & 87.08 & \textbf{86.67} & \textbf{82.50} & 9.54 & 8.93\\
\bottomrule
\end{tabularx}
\end{table}

\section{Experimental Results}
\label{sec:results}

After describing the setup (Section~\ref{sec:setup}), we ask what outcome
verification reveals beyond conversation ratings
(Section~\ref{sec:benchmark-results}), whether learning from verified
outcomes improves interaction (Section~\ref{sec:learning-results}),
whether the gains extend to other partners and tasks
(Section~\ref{sec:transfer}), and which components of \method{}
contribute (Section~\ref{sec:ablation}).

\subsection{Experimental Setup}
\label{sec:setup}

\paragraph{Training.}
We train Qwen3-8B \citep{yang2025qwen3} with LoRA \citep{hu2022lora}
against a frozen Qwen3-8B partner. Each RL run uses 384 training
conversations: one preassigned role per training scenario, balanced
across scenarios, with two samples per world. During training, Integrity
is scored by a frozen local Qwen3.8-27B judge \citep{qwen2026qwen3827b}.
To keep the number of sampled conversations equal across methods, the
group-based baselines (GRPO, Dr.GRPO, GiGPO, and GraphGPO), which draw
$G=4$ conversations per world, cover 48 training scenarios, whereas PPO
and \method{}, which draw two per world, cover all 96. \method{} uses $\alpha=0.25$ and $\beta=0.10$.
Appendix~\ref{app:sft-initialization} reports an additional SFT-initialization
study, and Appendix~\ref{app:training-cost} reports training costs.

\paragraph{Evaluation.}
The test set has 240 tasks (60 scenarios, two worlds, two roles), each
played against the frozen Qwen3-8B partner. Feasibility is computed by
rules; Integrity and the SOTOPIA ratings are judged by Qwen3.8-Max
\citep{alibaba2026qwen38max}, a different model from the training judge,
following rubric-based LLM evaluation \citep{zheng2023judging,liu2023geval}.
Failed generations count as failures. On 240 re-judged outputs,
Qwen3.8-Max and Kimi-K3 agree on 90.0\% of Integrity decisions (Cohen's
$\kappa=0.78$; Appendix~\ref{app:judge-agreement}).

\begin{table}[t]
\centering\small\setlength{\tabcolsep}{3pt}\renewcommand{\arraystretch}{1.08}
\caption{\textbf{Seven-system training comparison.} 240 test tasks per row; 384 training dialogues per RL method. To match this budget, group-based baselines ($G=4$) cover 48 training scenarios, and PPO and PaCPO cover 96. Goal and Believability (0--10) are not optimized. Best values are bold.}\label{tab:methods}
\begin{tabularx}{\linewidth}{@{}l*{5}{>{\raggedleft\arraybackslash}X}@{}}
\toprule
Method & \multicolumn{3}{c}{Verifiable outcomes (\%)} & \multicolumn{2}{c}{SOTOPIA ratings}\\
\cmidrule(lr){2-4}\cmidrule(l){5-6}
 & Feasibility & Integrity & \mbox{Ver. Succ.} & Goal & Believ.\\
\midrule
Base \tabref{yang2025qwen3} & 49.58 & 25.83 & 20.42 & 7.47 & 6.31\\
PPO \tabref{schulman2017ppo} & 50.00 & 33.75 & 25.00 & 7.58 & 6.51\\
\addlinespace[2pt]
GRPO \tabref{shao2024deepseekmath} & \textbf{52.92} & 31.25 & 26.25 & 7.83 & 6.62\\
\addlinespace[2pt]
Dr.GRPO \tabref{liu2025drgrpo} & 50.42 & 27.92 & 21.25 & 7.67 & 6.39\\
\addlinespace[2pt]
GiGPO \tabref{feng2025gigpo} & 52.08 & 30.42 & 24.58 & \textbf{7.86} & 6.60\\
\addlinespace[2pt]
GraphGPO \tabref{cheng2026graphgpo} & 52.50 & 31.67 & 24.17 & 7.84 & 6.60\\
\addlinespace[2pt]
\rowcolor{qwenpurple!9}
PaCPO & \textbf{52.92} & \textbf{35.42} & \textbf{28.33} & 7.83 & \textbf{6.70}\\
\bottomrule
\end{tabularx}
\end{table}

\subsection{Multi-Model Benchmark}
\label{sec:benchmark-results}

Table~\ref{tab:benchmark} reports all thirteen models. Verified Success
ranges from 20.42\% to 82.50\%, leaving headroom even for the strongest
models. The two checks capture different failures: Qwen3-8B passes
Feasibility on 49.58\% of tasks but reaches only 20.42\% Verified Success,
mostly due to Integrity failures.

Conversation ratings and verified outcomes can rank models differently.
GPT-5.6 Luna scores above Gemini 3.7 Flash on Goal (9.61 vs.\ 9.54) but
far below it on Verified Success (52.08\% vs.\ 82.50\%). Pooled over all
models, 35.99\% of conversations with Goal $\geq8$ fail Verified Success
(Appendix~\ref{app:outcome-diagnostics}), because the recorded plan is
infeasible or the conversation does not support it.

Dialogue itself matters. On 12 development scenarios with hidden
conditions, dialogue raises Base Feasibility by 50 points over
submitting records directly
(Appendix~\ref{app:outcome-diagnostics}).

\begin{figure}[t]
\centering
\begin{minipage}[t]{0.416667\linewidth}
\centering\vspace{0pt}
\includegraphics[width=\linewidth]{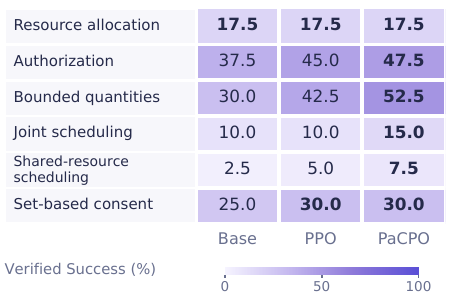}
\caption{\textbf{Verified Success by interaction family.} Forty test tasks per family with a fixed partner; bold marks the best score in each row.}\label{fig:family-joint-main}
\end{minipage}\hfill
\begin{minipage}[t]{0.5625\linewidth}
\centering\vspace{0pt}
\includegraphics[width=\linewidth,trim=0 14bp 0 0,clip]{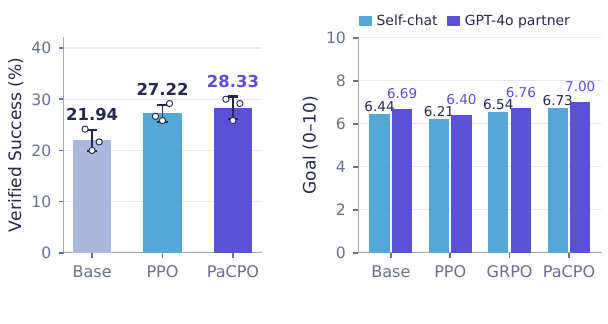}
\caption{\textbf{Self-play and SOTOPIA transfer.} Left: HICO Verified Success, mean $\pm$ SD over three generation repeats. Right: SOTOPIA Goal on 70 environment--character pairs per partner setting, judged by GPT-4o.}\label{fig:robustness-results}
\end{minipage}

\par\vspace{1em}
\begingroup
\makeatletter\def\@captype{table}\makeatother
\centering\small\setlength{\tabcolsep}{4pt}\renewcommand{\arraystretch}{1.12}
\caption{\textbf{External interaction tasks.} Task-specific training evaluated on 154 MultiWOZ and 86 CaSiNo test dialogues. All metrics are percentages; Norm. Nash denotes normalized Nash surplus.}
\label{tab:external-main}
\begin{tabularx}{\linewidth}{@{}l*{3}{>{\raggedleft\arraybackslash}X}@{}}
\toprule
& MultiWOZ & \multicolumn{2}{c}{CaSiNo}\\
\cmidrule(lr){2-2}\cmidrule(l){3-4}
Method & Comm. & Agreement & Norm. Nash\\
\midrule
Base \tabref{yang2025qwen3} & 79.87 & 36.05 & 27.40\\
PPO \tabref{schulman2017ppo} & 80.52 & 60.47 & 45.86\\
GRPO \tabref{shao2024deepseekmath} & 79.87 & 53.49 & 42.14\\
\rowcolor{qwenpurple!9}
\textbf{PaCPO (ours)} & \textbf{81.17} & \textbf{63.95} & \textbf{49.37}\\
\bottomrule
\end{tabularx}
\endgroup
\end{figure}

\subsection{Learning from Verified Outcomes}
\label{sec:learning-results}

Table~\ref{tab:methods} compares RL methods trained with the same
terminal reward and number of dialogues. \method{} achieves the highest
Verified Success, improving Base by 7.92 points to 28.33\%.
\method{} and GRPO reach the same
Feasibility, but \method{} has higher Integrity, so its gain comes from
conversations that better support feasible plans.

Although SOTOPIA ratings are not optimized, \method{} also improves Goal
and Believability over Base (Appendix~\ref{app:fullscores}). In a blinded
pilot with two expert annotators, human judgments agree with Feasibility
on 100\% of cases and with Qwen3.8-Max Integrity on 82.22\%; the
annotators also prefer \method{} over Base for helpfulness more often
than the reverse (8 vs.\ 5 of 24 pairs; Appendix~\ref{app:human-review}).

\paragraph{Where outcome learning helps.}
Gains vary by family (Figure~\ref{fig:family-joint-main}). The largest
improvements over Base are in bounded quantities (+22.50 points) and
authorization (+10.00 points). In bounded quantities, PPO and \method{}
reach the same Feasibility, so \method{}'s additional gain again comes
from Integrity. Resource allocation stays at 17.5\% Verified Success for
Base, PPO, and \method{}, and both scheduling families remain difficult
(Appendix~\ref{app:perfamily}).

\begin{figure}[t]
\centering
\includegraphics[width=\linewidth]{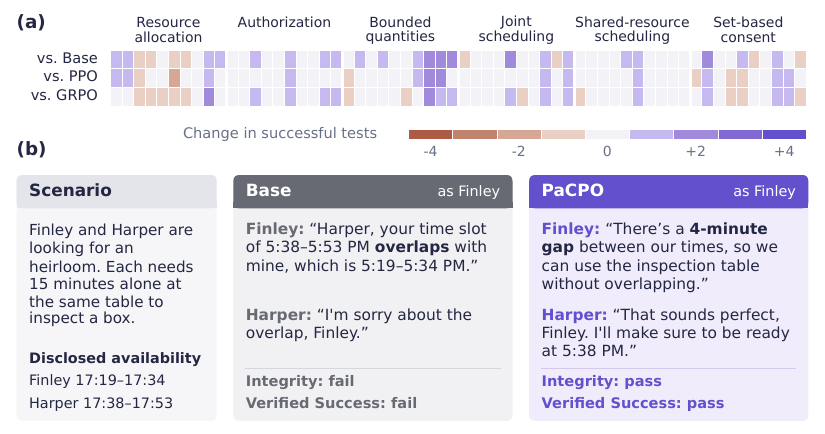}
\caption{\textbf{Scenario-level changes and a dialogue example.}
(a) PaCPO's change in successful tests relative to each baseline across
60 scenarios (four tests per column: two worlds and two roles).
(b) Verbatim excerpts with the same availability information and Finley
as the tested role. Both runs submit the same feasible records, but only
PaCPO's dialogue supports them.}
\label{fig:paired-dialogue}
\end{figure}

\paragraph{What changes in the conversation?}
Figure~\ref{fig:paired-dialogue}a breaks down the aggregate gains into
changes across individual scenarios. Improvements and regressions occur
within the same family, including the difficult scheduling settings.
Figure~\ref{fig:paired-dialogue}b shows one such improvement. Finley and
Harper disclose nonoverlapping windows for using a shared table. Base
incorrectly calls them overlapping, and its partner accepts that error.
\method{} instead identifies the four-minute gap and obtains agreement.
Both runs ultimately submit the same feasible records, but only
\method{}'s conversation supports them. This example illustrates why
checking the final records alone misses an important part of successful
coordination: the participants must also establish the plan in dialogue.

\subsection{Self-Play and Interactive Transfer}
\label{sec:transfer}
\label{sec:robustness}
We also evaluate self-play, transfer to SOTOPIA, and training on other interaction tasks.

\paragraph{Self-play.}
Self-play replaces the frozen partner with another copy of the evaluated checkpoint, testing robustness when both participants change. Across three generation repeats, PaCPO reaches 28.33\% mean Verified Success, exceeding Base by 6.39 points and PPO by 1.11 points (Figure~\ref{fig:robustness-results}, left). Gains persist in this setting, with a smaller margin over PPO than against the fixed partner. Full results appear in Appendix~\ref{app:interactive}.

\paragraph{SOTOPIA transfer.}
SOTOPIA \citep{zhou2023sotopia} tests direct transfer. Using iStar's partner settings and GPT-4o scoring \citep{liu2025istar}, we evaluate HICO checkpoints on 70 environment--character pairs without further training. PaCPO leads in both partner settings (Figure~\ref{fig:robustness-results}, right), connecting outcome training to social performance beyond HICO's checks (Appendix~\ref{app:interactive}).

\paragraph{Training on other interaction tasks.}
MultiWOZ \citep{budzianowski2018multiwoz} constraint communication and CaSiNo \citep{chawla2021casino} negotiation test task-specific training (Table~\ref{tab:external-main}; Appendix~\ref{app:external-training}). PaCPO gives a small communication gain on MultiWOZ and leads agreement and normalized Nash surplus on CaSiNo. Gains are larger in bargaining than in constraint communication.

\subsection{Credit Ablations}
\label{sec:ablation}
\begin{table}[t]
\centering\small\setlength{\tabcolsep}{4pt}\renewcommand{\arraystretch}{1.08}
\caption{\textbf{PaCPO credit ablations.} 240 test tasks per condition; all values are percentages.}\label{tab:controls}
\begin{tabularx}{\linewidth}{@{}l*{3}{>{\raggedleft\arraybackslash}X}@{}}
\toprule
Training condition & Feasibility & Integrity & \mbox{Ver. Succ.}\\
\midrule
Same-world candidate control & 50.83 & 32.08 & 26.25\\
Shuffled credit positions & 50.42 & 32.92 & 26.25\\
Without disclosure gate & 49.58 & 33.33 & 25.83\\
\rowcolor{qwenpurple!9}
PaCPO (full method) & \textbf{52.92} & \textbf{35.42} & \textbf{28.33}\\
\bottomrule
\end{tabularx}
\end{table}
Table~\ref{tab:controls} isolates three design choices. Replacing the
other world's plan with a second plan from the same world tests whether
the paired worlds matter, and shuffling credit across turns tests whether
its timing matters. Both reduce Verified Success to 26.25\%, 2.08 points
below full \method{} (Appendix~\ref{app:implementation}). Opening the
gate before the differing condition is disclosed reduces Verified Success
to 25.83\%, the largest drop, consistent with assigning credit only
after disclosure.

During training, paired credit is nonzero at 55.6\% of gate-open turns
(Appendix~\ref{app:perfamily}); otherwise,
the two worlds' plans are equally feasible, and only the within-world and
terminal terms apply.

\section{Conclusion}
\label{sec:conclusion}

Effective social interaction should be judged not only by the
conversation but also by the plan it produces. \bench{} evaluates this
with paired scenarios and two checks, Feasibility and Integrity, whose
conjunction defines Verified Success. Across thirteen LLMs, high
conversation ratings often accompany plans that fail these checks.
\method{} uses the paired worlds to turn end-of-conversation outcomes into
turn-level credit, improving Verified Success over the base model and RL
baselines. These results suggest that verified outcomes complement
conversation ratings, both for evaluation and as a training signal. The
benchmark currently covers six coordination families with structured
outcomes, and Integrity relies on an LLM judge; extending verification
to less structured interactions and expanding human validation are
important next steps (Appendices~\ref{app:limitations}
and~\ref{app:human-validation}).

\subsection*{AI use statement}
Generative AI assisted research design and analysis, data and code development, and manuscript preparation, including figures. Model-based data generation and evaluation are described in the relevant sections. The authors take full responsibility for the final work.

\subsection*{Ethics statement}
The scenarios use fictional or adapted role-play material. Their operational conditions concern feasible coordination and do not establish judgments about real individuals. Broader social quality remains distinct from feasibility under these formal conditions. Release preparation includes provenance and applicable licenses; released artifacts should exclude identifying information.

\subsection*{Reproducibility statement}
Appendices~\ref{app:families} and~\ref{app:metrics} describe the scenario families and evaluation criteria; Appendix~\ref{app:implementation} gives the training configuration and algorithm, and Appendices~\ref{app:interactive} and~\ref{app:external-training} document the external evaluation protocols. Tables and figures derive from recorded results. The intended release includes scenario lineage, role cards, verifier tests, prompts, rubrics, training settings, and checkpoint IDs.

\bibliography{references}
\bibliographystyle{colm2024_conference}
\clearpage
\appendix
\pdfbookmark[0]{Appendix Contents}{appendix-contents}
\section*{Appendix Contents}
\begingroup
\newcommand{\appendixentry}[1]{%
  \noindent\hyperref[#1]{\makebox[2em][l]{\ref*{#1}}\nameref*{#1}}%
  \nobreak\dotfill\hyperref[#1]{\pageref*{#1}}\par\medskip
}
\appendixentry{app:families}
\appendixentry{app:metrics}
\appendixentry{app:fullscores}
\appendixentry{app:benchmark}
\appendixentry{app:implementation}
\appendixentry{app:realcase}
\appendixentry{app:outcome-diagnostics}
\appendixentry{app:interactive}
\appendixentry{app:external-training}
\appendixentry{app:perfamily}
\appendixentry{app:human-validation}
\appendixentry{app:limitations}
\endgroup
\clearpage
\section{Scenario Families}
\label{app:families}
\begin{figure}[!htb]
\centering
\includegraphics[width=\linewidth]{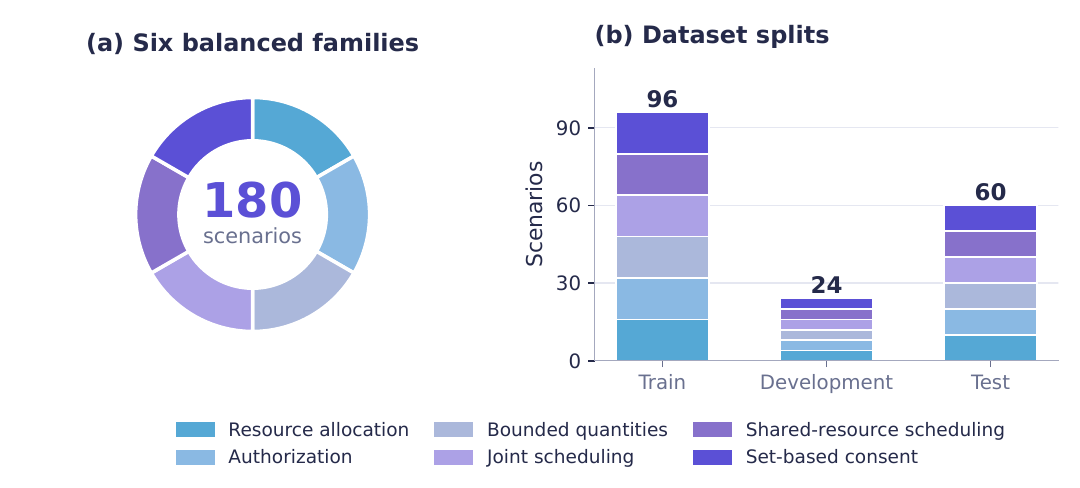}
\caption{\textbf{Dataset composition.} Each family contributes 30 scenarios: 16 training, 4 development, and 10 test. Two worlds per scenario yield 360 worlds; role-wise evaluation yields 96 development and 240 test tasks.}\label{fig:data-composition}
\end{figure}

HICO-Bench groups scenarios by the coordination requirement that determines which arrangements are feasible. These requirements apply across settings such as work, family, and commerce.

Table~\ref{tab:families} gives examples and verification rules; Figure~\ref{fig:data-composition} shows the balanced distribution. Each family has 30 scenarios: 16 training, 4 development, and 10 test, with two worlds per scenario. Shared-resource scheduling verifies two non-overlapping reservations. Set-based consent checks explicit required and allowed item sets.

Feasibility requires matching independent records and the correct outcome kind: agreement when a feasible plan exists, decline when none exists, or defer when the request requires authorization that neither role has.

\section{Evaluation Metrics}
\label{app:metrics}

In addition to the primary outcomes and SOTOPIA ratings, we report three
exploratory diagnostics: \emph{Information-gap Management} (IM),
\emph{Grounded Understanding Accuracy} (GU), and
\emph{Information-informed Action Appropriateness} (IA). These use the
frozen task-specific rubric and are reported separately, outside the
primary metrics and training reward.

\begin{table}[!htb]
\centering\small
\caption{\textbf{Evaluation metrics.} Binary outcomes and social dimensions with their scoring mechanisms. The original seven dimensions retain SOTOPIA ranges \citep{zhou2023sotopia}.}
\begin{tabularx}{\linewidth}{@{}lclY@{}}
\toprule
Metric & Range & Mechanism & Interpretation\\
\midrule
Feasibility & $\{0,1\}$ & Rules & Compatible records with terms satisfying the modeled conditions.\\
Integrity & $\{0,1\}$ & Language judge & No unresolved factual error, loop, commitment mismatch, or protocol leakage.\\
Verified Success & $\{0,1\}$ & Product & Feasibility and Integrity both pass.\\
\midrule
Goal & $[0,10]$ & Language judge & Attainment of the tested role's goal.\\
Knowledge & $[0,10]$ & Language judge & Acquisition of relevant information.\\
Believability & $[0,10]$ & Language judge & Naturalness and role consistency.\\
Relationship & $[-5,5]$ & Language judge & Effect on the relationship.\\
Secret & $[-10,0]$ & Language judge & Preservation of protected information.\\
Social Rules & $[-10,0]$ & Language judge & Social or legal violations.\\
Material Benefits & $[-5,5]$ & Language judge & Financial and material effects.\\
\midrule
IM & $[0,10]$ & Language judge & Information gaps resolved or appropriately kept conditional.\\
GU & $[0,10]$ & Language judge & Accurate interpretation supported by available evidence.\\
IA & $[0,10]$ & Language judge & Sufficiently specified responses using available information.\\
\bottomrule
\end{tabularx}
\end{table}

IM does not require a question when the necessary information is already available.
GU distinguishes stated facts from assumptions and tracks correction.
IA includes proposals, conditional commitments, deferral, and refusal rather than physical action.
IA diagnoses information-conditioned response quality.

\begin{table}[!htb]
\centering\small\setlength{\tabcolsep}{4pt}\renewcommand{\arraystretch}{1.12}
\caption{\textbf{Anchors for the three diagnostic ratings.} Each dimension evaluates the tested role over the complete dialogue using the frozen rubric.}
\label{tab:three-anchors}
\begin{tabularx}{\linewidth}{@{}lYYY@{}}
\toprule
Score & IM: information gaps & GU: interpretation & IA: response appropriateness\\
\midrule
0--2 & Commits despite a critical manageable unknown, or inquiry seriously blocks progress. & Material fabrication or persistent misunderstanding of facts, roles, or consent. & An unretracted response violates a critical known condition or invents authority.\\
\addlinespace[3pt]
3--5 & Useful exchange occurs, but a consequential gap remains unmanaged. & Only generic acknowledgment, or an unresolved material interpretive error. & A useful direction, but an essential term is missing or unsupported despite an opportunity.\\
\addlinespace[3pt]
6--8 & Main uncertainty handled, with a specific omission, redundancy, or delay. & Main facts understood, with a limited ambiguity or relevant corrected error. & Mainly feasible and concrete, with a limited coordination or scope weakness.\\
\addlinespace[3pt]
9--10 & Demonstrated sufficient information, or a genuinely conditional step managing the dependency. & Concrete responses demonstrate sufficient, accurate understanding and preserve uncertainty. & A sufficiently specified, grounded response; justified refusal or referral can qualify.\\
\bottomrule
\end{tabularx}
\end{table}

Each rating includes a short explanation linked to dialogue turn IDs. The anchors assess how the tested role manages information, interprets disclosures, and develops a response. Scores of 9--10 require concrete dialogue evidence that the relevant information has been handled sufficiently and accurately.

\FloatBarrier
\section{Additional Training Results}
\label{app:fullscores}

\subsection{SFT Initialization}
\label{app:sft-initialization}
Table~\ref{tab:sft} compares Base, SFT, PaCPO, and SFT followed by PaCPO (SFT + PaCPO) to examine how demonstration-based initialization complements outcome learning. This four-condition study supplements the Base-initialized algorithm comparison in Table~\ref{tab:methods}.

Applying \method{} after SFT raises Verified Success from 25.83\% (SFT
alone) to 31.67\%, the highest in our experiments
(Table~\ref{tab:sft}). The gain comes mainly from Integrity: the
Feasibility of SFT-initialized \method{} (48.75\%) remains slightly below
that of Base, and its Goal is lower than that of Base-initialized
\method{}.

\begin{table}[!htb]
\centering\small\setlength{\tabcolsep}{4pt}\renewcommand{\arraystretch}{1.08}
\caption{\textbf{SFT initialization and subsequent PaCPO.} 240 test tasks per condition. SFT uses 209 accepted teacher dialogues (991 examples, 64 updates); each RL stage uses 384 dialogues. All generated failures remain in the outcome denominator.}\label{tab:sft}
\begin{tabularx}{\linewidth}{@{}l*{3}{>{\raggedleft\arraybackslash}X}@{}}
\toprule
Method & Feasibility & Integrity & \mbox{Ver. Succ.}\\
\midrule
Base \tabref{yang2025qwen3} & 49.58 & 25.83 & 20.42\\
SFT & 46.25 & 29.58 & 25.83\\
\rowcolor{qwenpurple!9}
PaCPO & \textbf{52.92} & 35.42 & 28.33\\
\rowcolor{qwenpurple!9}
SFT + PaCPO & 48.75 & \textbf{36.25} & \textbf{31.67}\\
\bottomrule
\end{tabularx}
\end{table}

\subsection{Complete Social-Quality Scores}
\label{app:complete-social}
Table~\ref{tab:full-social} brings together all training conditions. The first seven score columns are SOTOPIA quality checks; the final three are exploratory process diagnostics. Dimensions retain their original scales; zero on Social Rules denotes no penalty. PaCPO improves Goal and Believability over Base, while SFT-initialized PaCPO obtains the highest Believability. The differing profiles complement the outcome comparison in Table~\ref{tab:methods}.
\begin{table}[!htb]
\centering\scriptsize\setlength{\tabcolsep}{2pt}\renewcommand{\arraystretch}{1.18}
\caption{\textbf{Complete SOTOPIA ratings for training methods.} Seven SOTOPIA dimensions and three process diagnostics, reported individually on their original scales. Know., Believ., Rel., and Mat. denote Knowledge, Believability, Relationship, and Material Benefits; Rules denotes Social Rules. All columns are higher-is-better; bold marks column-best values. SFT uses 239 scored dialogues; the other conditions use 240.}\label{tab:full-social}
\begin{tabularx}{\linewidth}{@{}l*{10}{>{\raggedleft\arraybackslash}X}@{}}
\toprule
Method & \multicolumn{7}{c}{SOTOPIA ratings} & \multicolumn{3}{c}{Diagnostics}\\
\cmidrule(lr){2-8}\cmidrule(l){9-11}
 & Goal & Know. & Believ. & Rel. & Secret & Rules & Mat. & IM & GU & IA\\
Range & $[0,10]$ & $[0,10]$ & $[0,10]$ & $[-5,5]$ & $[-10,0]$ & $[-10,0]$ & $[-5,5]$ & $[0,10]$ & $[0,10]$ & $[0,10]$\\
\midrule
Base & 7.47 & 7.41 & 6.31 & 2.15 & -1.43 & \textbf{0.00} & 1.08 & 7.87 & 7.56 & 7.09\\
PPO & 7.58 & 7.61 & 6.51 & \textbf{2.28} & \textbf{-1.24} & \textbf{0.00} & 1.20 & 8.12 & 7.93 & 7.51\\
GRPO, $G=4$ & 7.83 & 7.65 & 6.62 & 2.24 & -1.38 & \textbf{0.00} & 1.28 & 8.13 & 7.89 & 7.50\\
Dr.GRPO, $G=4$ & 7.67 & 7.67 & 6.39 & 2.25 & -1.35 & \textbf{0.00} & 1.28 & 8.16 & 7.97 & 7.49\\
GiGPO, $G=4$ & \textbf{7.86} & \textbf{7.72} & 6.60 & 2.25 & -1.37 & -0.03 & 1.23 & 8.15 & 7.89 & 7.54\\
GraphGPO, $G=4$ & 7.84 & 7.70 & 6.60 & 2.18 & -1.45 & -0.02 & \textbf{1.31} & \textbf{8.25} & \textbf{8.01} & \textbf{7.61}\\
\rowcolor{qwenpurple!9}
PaCPO & 7.83 & 7.69 & 6.70 & 2.19 & -1.42 & -0.02 & 1.21 & 8.18 & 7.95 & 7.57\\
\midrule
\multicolumn{11}{l}{\emph{Demonstration-based initialization}}\\
SFT & 7.35 & 7.57 & 6.62 & 2.12 & -1.48 & -0.09 & 1.14 & 7.66 & 7.10 & 6.89\\
\rowcolor{qwenpurple!9}
SFT + PaCPO & 7.29 & 7.42 & \textbf{7.03} & 1.97 & -1.41 & -0.08 & 1.18 & 7.72 & 7.47 & 7.17\\
\bottomrule
\end{tabularx}
\end{table}

\section{Model Benchmark}
\label{app:benchmark}
\begin{figure}[!htb]
\centering
\includegraphics[width=\linewidth]{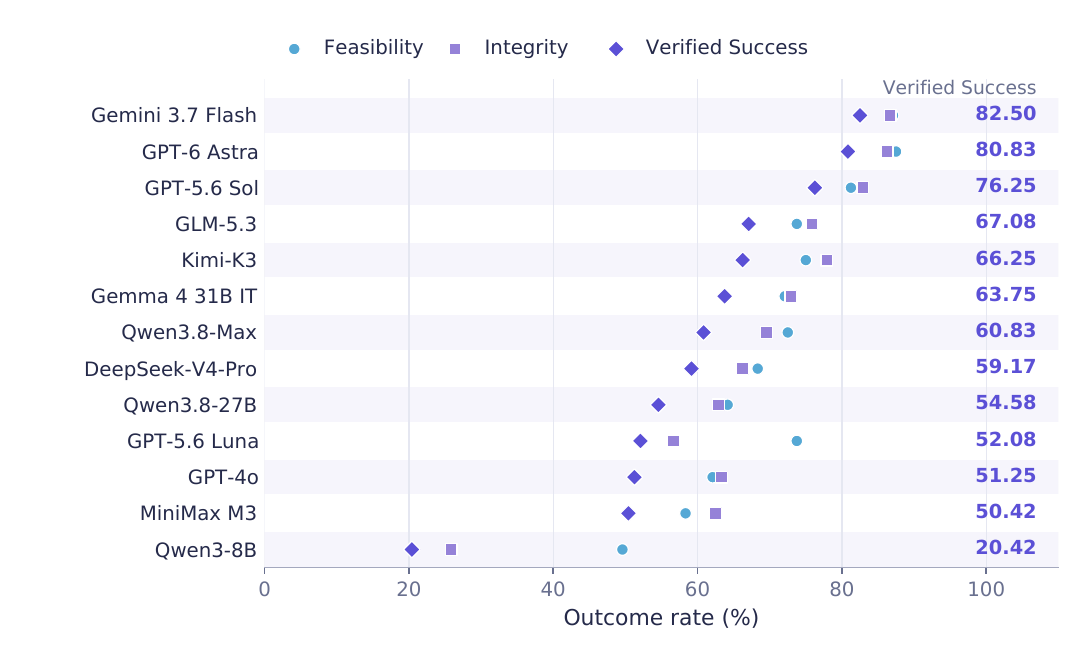}
\caption{\textbf{Verified outcomes across thirteen models.} Each point summarizes 240 tasks; right-hand labels give Verified Success. Complete SOTOPIA ratings appear in Table~\ref{tab:bench-social}.}\label{fig:benchmark-overview}
\end{figure}
\begin{table}[!htb]
\centering\scriptsize\setlength{\tabcolsep}{2pt}\renewcommand{\arraystretch}{1.18}
\caption{\textbf{Complete SOTOPIA ratings for thirteen models.} Model order follows Table~\ref{tab:benchmark}; abbreviations and score ranges follow Table~\ref{tab:full-social}. Bold marks column-best values. Luna and Sol each have 239 scored dialogues; the other models have 240. All 240 attempts enter the binary-outcome denominators.}\label{tab:bench-social}
\begin{tabularx}{\linewidth}{@{}l*{10}{>{\raggedleft\arraybackslash}X}@{}}
\toprule
Model & \multicolumn{7}{c}{SOTOPIA ratings} & \multicolumn{3}{c}{Diagnostics}\\
\cmidrule(lr){2-8}\cmidrule(l){9-11}
 & Goal & Know. & Believ. & Rel. & Secret & Rules & Mat. & IM & GU & IA\\
Range & $[0,10]$ & $[0,10]$ & $[0,10]$ & $[-5,5]$ & $[-10,0]$ & $[-10,0]$ & $[-5,5]$ & $[0,10]$ & $[0,10]$ & $[0,10]$\\
\midrule
Qwen3-8B & 7.47 & 7.41 & 6.31 & 2.15 & -1.43 & \textbf{0.00} & 1.08 & 7.87 & 7.56 & 7.09\\
MiniMax M3 & 9.18 & 8.31 & 8.68 & 2.80 & -0.80 & -0.01 & 1.40 & 9.31 & 9.28 & 9.19\\
GPT-4o & 9.12 & 8.22 & 8.71 & 2.68 & -0.76 & \textbf{0.00} & 1.32 & 9.06 & 8.99 & 8.85\\
GPT-5.6 Luna & 9.61 & 8.47 & 8.63 & 2.38 & -0.67 & \textbf{0.00} & 1.31 & 9.55 & 9.62 & 9.52\\
Qwen3.8-27B & 9.21 & 8.30 & 8.72 & 2.60 & -0.84 & \textbf{0.00} & 1.28 & 9.12 & 9.07 & 9.03\\
DeepSeek-V4-Pro & 9.44 & 8.46 & 8.88 & 2.85 & -0.70 & \textbf{0.00} & 1.39 & 9.50 & 9.47 & 9.43\\
Qwen3.8-Max & 9.62 & 8.53 & 8.95 & \textbf{3.10} & -0.74 & \textbf{0.00} & 1.33 & 9.59 & 9.72 & 9.61\\
Gemma 4 31B IT & 9.38 & 8.21 & 8.80 & 2.65 & \textbf{-0.49} & \textbf{0.00} & 1.43 & 9.13 & 9.22 & 9.10\\
Kimi-K3 & 9.52 & 8.51 & 8.94 & 2.61 & -0.75 & \textbf{0.00} & 1.34 & 9.52 & 9.56 & 9.49\\
GLM-5.3 & 9.65 & 8.44 & 8.96 & 2.84 & -0.50 & \textbf{0.00} & \textbf{1.45} & 9.59 & 9.66 & 9.61\\
GPT-5.6 Sol & 9.63 & \textbf{8.60} & 8.95 & 1.99 & -0.65 & \textbf{0.00} & 1.33 & 9.74 & 9.81 & 9.79\\
GPT-6 Astra & \textbf{9.75} & \textbf{8.60} & \textbf{9.01} & 2.17 & -0.71 & \textbf{0.00} & 1.30 & \textbf{9.76} & \textbf{9.84} & \textbf{9.84}\\
Gemini 3.7 Flash & 9.54 & 8.54 & 8.93 & 2.22 & -0.66 & \textbf{0.00} & 1.26 & 9.67 & 9.74 & 9.75\\
\bottomrule
\end{tabularx}
\end{table}

Figure~\ref{fig:benchmark-overview} compares Feasibility, Integrity, and Verified Success for every model, and Table~\ref{tab:bench-social} reports their complete SOTOPIA ratings. Goal scores are high for most API models, while the outcome rates remain more widely separated. This contrast motivates evaluating conversation ratings and verified outcomes together.

\section{Training Details}
\label{app:implementation}

Algorithm~\ref{alg:pacpo} specifies the complete update corresponding to Figure~\ref{fig:pacpo-training}.
\begin{algorithm}[!htb]
\caption{Paired-Credit Policy Optimization (PaCPO)}
\label{alg:pacpo}
\small
\begin{algorithmic}[1]
\Require Paired scenarios $\mathcal D$; actor $\pi_\theta$; fixed partner $b$ and reference $\pi_{\mathrm{ref}}$;
\Statex \hspace{\algorithmicindent}value head $V_\psi$; checks $F,I,\mathcal V$; scales $\alpha=0.25$, $\beta=0.10$.
\For{each rollout batch}
  \State Freeze rollout policy $\pi_{\mathrm{old}}\gets\pi_\theta$ and value baseline $\bar V$; sample pairs from $\mathcal D$.
  \State Generate independent paired trajectories with $\pi_{\mathrm{old}}$ and $b$, including final records.
  \For{each trajectory pair $(\tau_0,\tau_1)$ with the same tested role}
    \State Compute $R_i\gets0.8F_i+0.2I_i$ for each world $i\in\{0,1\}$.
    \For{each world $i$ and actor dialogue prefix $t<T_i$}
      \State Sample $\widehat c_{i,t}\sim\pi_{\mathrm{old}}(\cdot\mid p,v_i^{\mathrm{act}},h_{i,t},\mathcal C)$, outside the dialogue.
      \State Set $\Phi_{i,t}\gets\alpha\mathcal V(w_i,\widehat c_{i,t})$; use zero for an invalid tentative plan.
    \EndFor
    \For{each matched actor-turn index $t$ with two valid tentative plans}
      \State Compute $M_{ij}^t\gets\mathcal V(w_i,\widehat c_{j,t})$ for $i,j\in\{0,1\}$.
      \State $g_t\gets\operatorname{Disjoint}(w_0,w_1)\land\operatorname{Known}(0,t)\land\operatorname{Known}(1,t)$.
      \State $\Phi_{i,t}\gets\Phi_{i,t}+\beta g_t(M_{ii}^t-M_{i,1-i}^t)$ for both worlds.
    \EndFor
    \For{each world $i$}
      \State Set $\Phi_{i,0}=\Phi_{i,T_i}=0$; $r_{i,T_i}=R_i$ and $r_{i,t<T_i}=0$.
      \State Redistribute $r'_{i,t}\gets r_{i,t}+\Phi_{i,t}-\Phi_{i,t-1}$; hence $\sum_t r'_{i,t}=R_i$.
      \State Place $r'_{i,t}$ on each output's last token; compute raw GAE $\widehat A$ using $\bar V$.
      \State Cache value targets $\widehat G\gets\widehat A+\bar V$ before advantage normalization.
    \EndFor
  \EndFor
  \State Normalize $\widehat A$ over rollout actor tokens to obtain $A$.
  \State Minimize the clipped actor loss with reference KL and the separate clipped value loss.
\EndFor
\State \Return $\pi_\theta$
\end{algorithmic}
\vspace{2pt}
{\footnotesize $T_i$ is the turn of the final record. PPO includes the tested role's dialogue and record tokens, not partner or tentative-plan tokens. $\operatorname{Known}(i,t)$ uses the tested role's card or an annotated disclosure at or before $t$; all cross-world checks remain training-side.}
\end{algorithm}

\begin{table}[!htb]
\centering
\small\setlength{\tabcolsep}{5pt}\renewcommand{\arraystretch}{1.12}
\caption{\textbf{Principal PaCPO configuration.}}
\begin{tabularx}{\linewidth}{@{}>{\raggedright\arraybackslash}p{.51\linewidth}Y@{}}
\toprule
Parameter & Value\\
\midrule
Policy / fixed partner & Qwen3-8B LoRA / frozen Qwen3-8B\\
Initialization & Base; no SFT adapter\\
Train scenarios / worlds / conversations & 96 / 192 / 384\\
Rollout batch / PPO epochs & 16 conversations / 2\\
Actor / value-head learning rate & $2\times10^{-6}$ / $2\times10^{-4}$\\
GAE discount / trace & $\gamma=1$ / $\lambda=0.95$\\
PPO ratio clip / value clip & 0.2 / 0.2\\
Reference KL / value-loss coefficient & 0.02 / 0.5\\
Training Integrity judge / reasoning & Qwen3.8-27B / low\\
Within-world / paired scale & $\alpha=0.25$ / $\beta=0.10$\\
Maximum output / alternating turns & 512 tokens / 20\\
Training / Test generation seed & 917 / 317917\\
Eligible / nonzero paired positions & 1,378 / 766\\
Accepted optimizer updates & 192\\
\bottomrule
\end{tabularx}
\end{table}

\paragraph{Training procedure.}
Both worlds of a scenario are sampled under the same behavior-policy version.
The two trajectories are not forced to have identical language or identical length.
After each turn of the tested role, PaCPO generates one tentative plan from the visible prefix. Tentative plans at matching turn indices are checked in both worlds. When one conversation lacks a corresponding turn or a valid tentative plan, the paired credit is zero, while within-world feasibility remains available.

Algorithm~\ref{alg:pacpo} gives the complete rollout update. Implementation details below specify value learning, token masking, and information boundaries.

\paragraph{Value learning.}
The policy has a separately optimized linear value head.
Its input hidden states are detached from the policy backbone. Proposal verification supplies process rewards, while the value head predicts return for the PPO baseline.
Each turn's redistributed reward is placed on the last actor token, and advantages are normalized over the rollout batch.
For a three-output illustration, let $R_i=1$ and $(\Phi_{i,0},\Phi_{i,1},\Phi_{i,2},\Phi_{i,3})=(0,0.25,0.35,0)$. The original reward sequence $(0,0,1)$ becomes $(0.25,0.10,0.65)$, retaining a total of $1$. This example illustrates the transformation; actual potentials depend on the tentative-plan checks and the gate.

\paragraph{Token-level advantages and targets.}
Within one trajectory, let $k=1,\ldots,K$ index the tested role's tokens, including the final record. The state $s_k$ contains the role-visible prefix and preceding tokens of the current output. Across outputs, the next actor state includes any intervening partner observation. Partner and tentative-plan tokens do not occupy positions in this reward sequence. Set $\widetilde r_k=r'_{i,t}$ on the final token of turn $t$, and zero elsewhere.

Let $\bar V_k$ be the cached, pre-update value prediction at $s_k$, with $\bar V_{K+1}=0$. Token-level GAE \citep{schulman2015gae} and the value target are
\begin{equation}
\begin{aligned}
\delta_k &= \widetilde r_k+\gamma\bar V_{k+1}-\bar V_k,\\
\widehat A_k &= \sum_{\ell=0}^{K-k}(\gamma\lambda)^\ell\delta_{k+\ell},
\qquad \widehat G_k=\widehat A_k+\bar V_k .
\end{aligned}
\label{eq:gae}
\end{equation}
We use $\gamma=1$ and $\lambda=0.95$. Cached values, raw advantages, and $\lambda$-return targets $\widehat G_k$ remain fixed during the two PPO epochs. The actor uses normalized advantages
\begin{equation}
A_k=\frac{\widehat A_k-\mu_{\mathcal R}}{\sigma_{\mathcal R}+10^{-8}},
\label{eq:advnorm}
\end{equation}
where $\mu_{\mathcal R}$ and $\sigma_{\mathcal R}$ are the mean and population standard deviation over all actor tokens in the 16-dialogue rollout batch $\mathcal R$. Value targets use raw, not normalized, advantages.

\paragraph{Actor loss and reference regularization.}
The implementation averages tokens within each dialogue, then averages dialogues within each four-dialogue optimizer minibatch $\mathcal B$. For a token quantity $f_{n,k}$ in dialogue $n$, define
\begin{equation}
\langle f\rangle_{\mathcal B}
=\frac{1}{|\mathcal B|}\sum_{n\in\mathcal B}\frac{1}{K_n}\sum_{k=1}^{K_n}f_{n,k}.
\label{eq:lossaverage}
\end{equation}
Suppressing the dialogue index, the PPO ratio and clipped actor loss \citep{schulman2017ppo} to \emph{minimize} are
\begin{equation}
\begin{aligned}
\rho_k&=\frac{\pi_\theta(a_k\mid s_k)}{\pi_{\mathrm{old}}(a_k\mid s_k)},\\
\mathcal L_{\mathrm{actor}}
&=-\left\langle\min\!\left(\rho_k A_k,
\operatorname{clip}(\rho_k,1-\epsilon,1+\epsilon)A_k\right)\right\rangle_{\mathcal B}
+\eta\mathcal L_{\mathrm{ref}},
\end{aligned}
\label{eq:ppo}
\end{equation}
with $\epsilon=0.2$ and $\eta=0.02$. We use the nonnegative sampled reference penalty from DeepSeekMath \citep[Eq.~4]{shao2024deepseekmath}:
\begin{equation}
\mathcal L_{\mathrm{ref}}=\left\langle e^{u_k}-u_k-1\right\rangle_{\mathcal B},
\qquad u_k=\log\pi_{\mathrm{ref}}(a_k\mid s_k)-\log\pi_\theta(a_k\mid s_k).
\label{eq:refkl}
\end{equation}
These quantities are evaluated on the stored rollout tokens. Current, behavior, and reference token probabilities are normalized over the same public-schema grammar support used for generation, including the end-of-sequence decision. KL regularization is added to the loss, not to $\widetilde r_k$ or the value target.

\paragraph{Clipped value regression.}
Let $V_k=V_\psi(s_k)$ be the current value prediction. Following the clipped value-regression implementation described by \citet{huang2022ppodetails}, with $\epsilon_V=0.2$, the value loss is
\begin{equation}
\begin{aligned}
V_k^{\mathrm{clip}}&=\bar V_k+\operatorname{clip}(V_k-\bar V_k,-\epsilon_V,\epsilon_V),\\
\mathcal L_V&=\frac{1}{2}\left\langle\max\!\left[
(V_k-\widehat G_k)^2,(V_k^{\mathrm{clip}}-\widehat G_k)^2
\right]\right\rangle_{\mathcal B}.
\end{aligned}
\label{eq:valueloss}
\end{equation}
The minimized loss is $\mathcal L=\mathcal L_{\mathrm{actor}}+0.5\mathcal L_V$. The value head receives detached policy hidden states, so value regression updates its own parameters but does not backpropagate through the policy backbone. The critic predicts $\lambda$-returns; the rule verifier supplies tentative-plan feasibility.

\paragraph{Information boundaries.}
Alternate-world state and undisclosed partner-private facts remain training-side.
They are never injected into actor or reference-policy prompts. Facts disclosed by the partner become ordinary dialogue context.
The hypothetical record is not treated as a demonstrated response.
When the differing condition is on the tested role's own card, it is available from the first turn. Otherwise, the frozen local judge receives both partner cards and the recorded dialogue to identify the earliest partner disclosure; turn indices determine which earlier tentative plans remain gated off. This annotation approximates when the differing condition becomes available to the tested role.

\paragraph{Comparability.}
The main methods share the backbone, partner, terminal reward, generation limits, and test protocol. They use 384 training dialogues; group-based baselines use $G=4$ and sample 48 scenarios, while PPO and PaCPO sample 96. PaCPO additionally generates tentative plans and applies cross-world rule checks. Comparisons match dialogue counts, not scenario coverage or total compute. Main comparisons use one training seed (917) and one test generation seed (317917).

\paragraph{Mechanism controls.}
\label{app:credit-controls}
The same-world control draws comparison candidates under the same world conditions. Shuffled credit positions preserve the computed credit values while changing their temporal placement. The disclosure-gate control retains the same disclosure annotations, validity checks, and disjointness condition, and applies paired credit regardless of whether the differing condition has appeared in both conversations. Table~\ref{tab:controls} reports all three comparisons.

\paragraph{Evaluation protocol.}
The main comparison, mechanism controls, and self-play use the same frozen Integrity rubric with Qwen3.8-Max. The seven SOTOPIA and three diagnostic rubrics are unchanged.

\subsection{GiGPO and GraphGPO Adaptation}
\label{app:group-baselines}

The GiGPO and GraphGPO rows in Table~\ref{tab:methods} use their released advantage-computation routines within the HICO dialogue trainer \citep{feng2025gigpo,cheng2026graphgpo}. We load the numerical function bodies unchanged from a pinned \texttt{verl-agent} checkout; the environment interface, observation representation, rollout schedule, and policy-update loop use our HICO implementation.\footnote{\raggedright\href{https://github.com/langfengQ/verl-agent/tree/20bd331bdbc9026a5668e11362178e10ab7400c8}{Source revision: \texttt{20bd331bdbc9}}. GiGPO uses \texttt{gigpo/core\_gigpo.py}; GraphGPO uses \texttt{recipe/GraphGPO/core\_graph.py}.}

\paragraph{Rollout groups and task feedback.}
Each group contains $G=4$ independent dialogues for the same scenario, world, and tested role, against the frozen Qwen3-8B partner. Each turn of the tested role, including the final record, is one step. The terminal reward $R$ is assigned to the final step; earlier rewards are zero. The collection budget is 384 dialogues: 48 scenarios $\times$ two worlds $\times$ four samples, with balanced interaction families and preassigned roles. Both methods are critic-free and generate no tentative plans.

\paragraph{Anchor-state representation.}
An anchor key hashes the exact actor-input token sequence together with a flag distinguishing dialogue turns from final records. It therefore contains the actor's role card, public instructions, and full visible history. Matching is restricted to the same scenario, world, and role. We disable approximate string matching and semantic state merging; partner-private fields and verifier state are not added to the key. This keeps decision-relevant history intact while making shared anchors depend on identical visible prefixes.

\paragraph{GiGPO advantages.}
For a dialogue ending at step $T$, GiGPO computes $U_t=\sum_{u=t}^{T}0.95^{u-t}r_u$, then compares these discounted returns among occurrences of the same anchor. The resulting step advantage is added to an episode-level advantage computed from $R$, with both coefficients equal to one. We use the released \texttt{mean\_std\_norm} mode: subtract the group mean and divide by the sample standard deviation plus $10^{-6}$. Episode normalization retains the upstream cross-step default, repeating each dialogue's $R$ at its steps when computing the statistics. Step normalization instead uses the corresponding anchor group; a singleton anchor has zero centered step advantage.

\paragraph{GraphGPO advantages.}
GraphGPO builds a directed graph separately for each rollout group. Nodes are the same exact-history anchors, and unit-weight edges connect successive actor steps. A dialogue with $R=1$ connects to a shared success node; other terminal nodes remain trajectory-specific. The released routine computes shortest-path distances to success. With distance normalization disabled, the step score is $10(0.10)^{D(s_{t+1})}$, where $D$ is the successor's distance to success. Unreachable nodes receive the upstream fallback distance, one greater than the largest finite distance. These scores are normalized within anchor groups and added to the same episode-level advantage, again with unit coefficients. This path score determines credit; it does not replace the terminal reward.

\paragraph{Policy update and scope.}
Each computed step advantage is broadcast to the tokens of that turn. Both adapters use the clipped actor loss and reference penalty above, with clipping 0.2, reference coefficient 0.02, actor learning rate $2\times10^{-6}$, and two optimization epochs. Tokens are averaged within dialogues before dialogues are averaged within four-dialogue optimizer minibatches. No further rollout-wide advantage normalization is applied, and partner tokens remain context only.

Exact-history grouping supports local comparisons only at repeated anchors. Once dialogue histories diverge, unmatched anchors retain the episode-level component without a local step comparison. The reported results characterize this exact-history HICO adaptation; the grouping and state representation are part of the comparison, alongside the advantage estimator.

\subsection{Training Cost}
\label{app:training-cost}

Table~\ref{tab:training-cost} accounts for both dialogue participants, tentative-plan generation, disclosure annotation, and optimizer updates. Counts include generated outputs later excluded from training. All three runs sample 384 dialogues; PPO and PaCPO retain 384, while GRPO retains 376 after two four-dialogue groups encounter partner-generation failures. These correspond to the $G=4$ GRPO and PaCPO checkpoints in the main comparison.

\begin{table}[!htb]
\centering\small\setlength{\tabcolsep}{4pt}\renewcommand{\arraystretch}{1.15}
\caption{\textbf{Training-generation and annotation cost.} Total output includes tested-role, partner, and tentative-plan tokens. Tentative-plan output is a subset of this total; disclosure requests are additional to terminal Integrity scoring.}
\label{tab:training-cost}
\begin{tabularx}{\linewidth}{@{}l*{4}{>{\raggedleft\arraybackslash}X}@{}}
\toprule
Method & \shortstack[r]{Total output\\tokens} & \shortstack[r]{Probe output\\tokens} & \shortstack[r]{Disclosure\\requests} & \shortstack[r]{Optimizer\\updates}\\
\midrule
PPO & 276,293 & 0 & 0 & 192\\
GRPO ($G=4$) & 282,431 & 0 & 0 & 188\\
\rowcolor{qwenpurple!9}
PaCPO & 310,863 & 57,405 & 216 & 192\\
\bottomrule
\end{tabularx}
\end{table}

PaCPO generates 1,869 tentative plans, contributing 57,405 output tokens and 2,024,795 input tokens. Including both dialogue participants, its generated output is 12.51\% higher than PPO's. Input counts sum unpadded request lengths, including repeatedly processed prefixes; judge tokens and policy optimization are separate costs. The 216 disclosure annotations complement 168 dialogues in which the differing condition is on the tested role's own card.

The local training scorer records 391, 386, and 603 scoring or annotation attempts for PPO, GRPO, and PaCPO, respectively, including validation retries. PaCPO's total includes 29 cached attempts from its first 16 dialogues, reused from an interrupted run before policy updates. These are counted once, yielding 574 new scorer attempts in the resumed run. Test-time Qwen3.8-Max evaluation is outside this training account.

\section{A Paired Scenario}
\label{app:realcase}

The training scenario \texttt{branding\_dilemma} illustrates authorization.
Avery, a new business owner uncertain about a brand change, requests an extra concept-review block from consultant Blair.
The public context states that complimentary hours require delegated authority and that a pending referral is not approval.
It does not disclose Avery's requested duration or Blair's authority limit.

\begin{table}[!htb]
\centering\small
\setlength{\tabcolsep}{5pt}\renewcommand{\arraystretch}{1.12}
\caption{\textbf{Paired authorization scenario.} Only Blair's private authorization limit changes.}
\begin{tabularx}{\linewidth}{@{}>{\raggedright\arraybackslash}p{.35\linewidth}YY@{}}
\toprule
Field & World $w_0$ & World $w_1$\\
\midrule
Avery's request & 1 hour of extra review & 1 hour of extra review\\
Blair's delegated limit & At most 2 hours & At most 0 hours\\
Referral authority & Agency lead & Agency lead\\
Appropriate record kind & Agreement & Defer\\
Approval / referral & True / none & False / agency lead\\
\bottomrule
\end{tabularx}
\end{table}

Both records must retain the requested one-hour parameter and the operation \texttt{approve\_extra\_review}.
The first world admits current authorization; the second requires truthful referral rather than invented approval.
Integrity additionally checks whether the conversation supports the recorded plan.
The changed authority boundary determines whether approval or referral is feasible.

\section{Outcome Reliability and Information Diagnostics}
\label{app:outcome-diagnostics}
\begin{figure}[!htb]
\centering
\includegraphics[width=\linewidth]{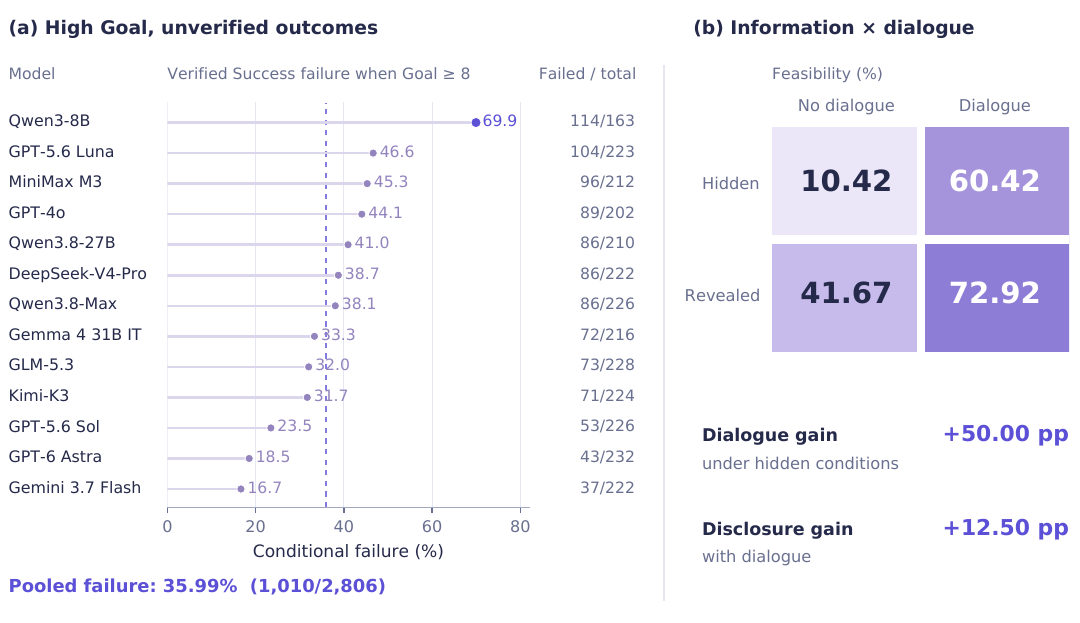}
\caption{\textbf{Why verify outcomes?} (a) Verified Success failure rate conditioned on Goal $\geq8$, with failed/eligible counts for each model. The dashed line is the pooled rate: 1,010/2,806 model--task observations (35.99\%). (b) Base under a $2\times2$ information-access and dialogue design: 12 development scenarios, 48 tasks per condition. Cells show rule-based Feasibility success; gains are percentage-point differences. Dialogue includes information exchange and negotiation.}\label{fig:task-evidence}
\end{figure}

\paragraph{High Goal does not certify valid outcomes.}
Among 163 Qwen3-8B conversations with Goal $\geq8$, 114 (69.94\%) fail Verified Success. Across all thirteen models, 1,010 of 2,806 high-Goal model--task observations fail (35.99\%; Figure~\ref{fig:task-evidence}a). These are repeated model--task observations, not distinct scenarios. A high role-goal rating does not certify a feasible, dialogue-supported outcome.

\paragraph{Information access and interaction.}
On 12 fixed development scenarios, Figure~\ref{fig:task-evidence}(b) separates information access from interaction. Without dialogue, revealing conditions raises Base Feasibility from 10.42\% to 41.67\%; with dialogue, it rises from 60.42\% to 72.92\%. Under hidden conditions, dialogue adds 50.00 points, combining information exchange and negotiation.

The diagnostic uses 12 fixed development scenarios, both worlds, and both tested roles, yielding 48 tasks per condition.
No-dialogue conditions have no text on which to judge Integrity or SOTOPIA ratings, so those fields are marked not applicable rather than zero.

\begin{table}[!htb]
\centering\small
\setlength{\tabcolsep}{4pt}\renewcommand{\arraystretch}{1.12}
\caption{\textbf{Information-access and dialogue diagnostic.} Conditions in Figure~\ref{fig:task-evidence}(b); all values are percentages. Own-record feasibility checks the tested role's record against the full scenario constraints, including both participants' conditions; Feasibility additionally requires both records to match. Both checks are rule-based. Verified Success here uses the original local Qwen3.8-27B judge, not Qwen3.8-Max.}
\begin{tabularx}{\linewidth}{@{}l*{3}{>{\raggedleft\arraybackslash}X}@{}}
\toprule
Condition & Feasibility & \shortstack[r]{Own-record\\feasibility} & \mbox{Ver. Succ.}\\
\midrule
No dialogue & 10.42 & 25.00 & --\\
Revealed + no dialogue & 41.67 & 52.08 & --\\
Hidden + dialogue & 60.42 & 62.50 & 31.25\\
Revealed + dialogue & 72.92 & 75.00 & 39.58\\
\bottomrule
\end{tabularx}
\end{table}

No dialogue removes information exchange and coordination. Revealing conditions restores the information while leaving agents to submit independently.

\section{Interactive Transfer and Self-Play}
\label{app:interactive}

\paragraph{SOTOPIA protocol.}
Figure~\ref{fig:robustness-results} (right) and Section~\ref{sec:robustness} report direct transfer of HICO checkpoints without further training. We evaluate 70 environment--character pairs per partner setting, sampled from 14 environments with a fixed selection seed. The two settings use the same checkpoint for both roles or GPT-4o as the partner. SOTOPIA \citep{zhou2023sotopia} uses iStar-style public prompts \citep{liu2025istar} and GPT-4o seven-dimensional scoring; the reported Goal is the tested Agent-1's score on the original 0--10 scale. Each partner setting uses one generation seed.

\paragraph{Complete self-play results.}
Table~\ref{tab:selfplay} supplements the Verified Success comparison in Figure~\ref{fig:robustness-results} (left) with Feasibility, Integrity, and standard deviations across three generation seeds. Both participants use the same checkpoint; each seed comprises 120 dialogues and 240 scored role views.
\begin{table}[!htb]
\centering\small\setlength{\tabcolsep}{4pt}\renewcommand{\arraystretch}{1.08}
\caption{\textbf{Full self-play results.} Both participants use the same checkpoint. Mean $\pm$ sample SD over three generation seeds; 120 dialogues and 240 scored role views per seed. Rates are percentages; generation failures remain in the denominator.}\label{tab:selfplay}
\begin{tabularx}{\linewidth}{@{}l*{3}{>{\raggedleft\arraybackslash}X}@{}}
\toprule
Method & Feasibility & Integrity & \mbox{Ver. Succ.}\\
\midrule
Base \tabref{yang2025qwen3} & 48.89 $\pm$ 2.68 & 29.17 $\pm$ 2.20 & 21.94 $\pm$ 2.10\\
PPO \tabref{schulman2017ppo} & 48.89 $\pm$ 2.55 & 31.67 $\pm$ 2.20 & 27.22 $\pm$ 1.73\\
\rowcolor{qwenpurple!9}
PaCPO & 51.94 $\pm$ 2.68 & 35.83 $\pm$ 2.50 & 28.33 $\pm$ 2.20\\
\bottomrule
\end{tabularx}
\end{table}

\section{External-Task Training}
\label{app:external-training}
We evaluate task-specific training in two settings: database-grounded recommendation and bilateral negotiation. Table~\ref{tab:external-main} reports their results.

\paragraph{MultiWOZ: eliciting constraints for a recommendation.}
We select hotel and restaurant goals from MultiWOZ 2.2 dialogue-state annotations \citep{zang2020multiwoz22} and use the corresponding MultiWOZ 2.1 entity databases \citep{eric2020multiwoz21}. Each source contributes one domain-specific recommendation task with explicit user constraints, such as area, price range, cuisine, or hotel amenities. We select goals with a database match and omit booking requests and requests for a preselected named entity. Train, development, and test retain their original source splits; repeated constraint profiles are removed across splits.

Each pair retains the public request and database while changing one private user constraint. Both worlds have valid recommendations, and their valid entity sets are disjoint. The visitor holds the requirements privately; the assistant asks questions, searches the database, and recommends one entity by name. A successful episode requires that the selected entity satisfy every private constraint and be unambiguously identified in the final message. This \emph{constraint communication} endpoint tests whether elicited requirements reach the actual recommendation. The test set contains 77 paired scenarios and 154 dialogues.

\paragraph{CaSiNo: negotiating under private preferences.}
We retain CaSiNo's campsite setting \citep{chawla2021casino}, with three units each of food, water, and firewood. Each participant has private high-, medium-, and low-priority items, valued at 5, 4, and 3 points per unit, together with reasons for those preferences. To construct paired worlds, we keep the tested participant's card fixed and replace the partner's complete preference-and-reason card with one from the same original split. This changes which allocations benefit the partner while retaining the public negotiation setting.

The participants negotiate a complete allocation that both accept. We report agreement rate and normalized Nash surplus, based on the Nash bargaining product \citep{nash1950bargaining}. We take the product of both participants' nonnegative utility gains above the five-point walk-away value and divide it by the maximum attainable product in that world. Episodes without agreement receive zero surplus. These measures distinguish reaching an agreement from finding a mutually beneficial allocation. The test set contains 43 paired scenarios and 86 dialogues.

\section{Family-Level Analysis}
\label{app:perfamily}
\begin{figure}[!htb]
\centering
\includegraphics[width=\linewidth]{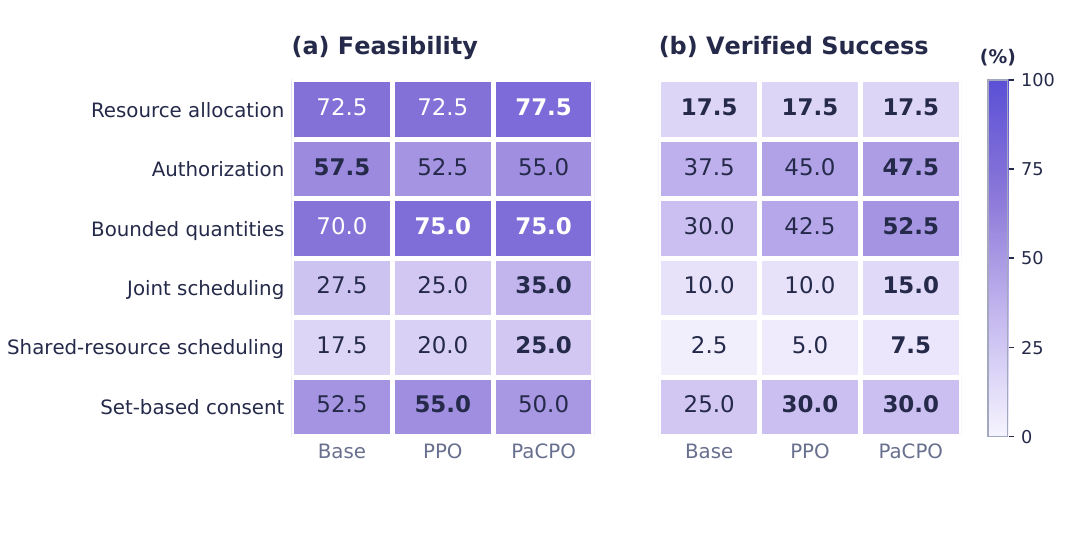}
\caption{\textbf{Results by interaction family.} Each cell contains 40 test tasks (ten scenarios, both worlds and roles). Both panels use the same color scale; bold marks the best method within each row and panel. Scheduling remains difficult across methods.}\label{fig:family-results}
\end{figure}

Figure~\ref{fig:family-results} reports Feasibility and Verified Success on 40 tasks per family: ten scenarios, two worlds, and two roles. All 60 test scenarios have disjoint valid outcome sets across their paired worlds.

\paragraph{Where outcome learning helps.}
PaCPO's largest Verified Success gain over Base occurs for bounded quantities, rising from 30.0\% to 52.5\%. It also improves authorization from 37.5\% to 47.5\%. For bounded quantities, PPO and PaCPO have identical Feasibility (75.0\%), but PaCPO has higher Verified Success (52.5\% versus 42.5\%). The additional gain therefore lies in jointly satisfying feasibility and dialogue integrity, rather than increasing feasibility alone.

\paragraph{What remains difficult.}
Resource allocation illustrates the remaining gap: PaCPO reaches 77.5\% Feasibility, yet Verified Success stays at 17.5\%, the same as Base and PPO. Producing a feasible split is thus easier than maintaining dialogue support for it. Joint scheduling and shared-resource scheduling remain the lowest-scoring PaCPO families, at 15.0\% and 7.5\% Verified Success. These tasks require compatible time commitments across participants, and their low Feasibility rates indicate that feasible coordination itself remains difficult.

Table~\ref{tab:perfamily-credit} complements these outcomes with the frequency of paired-credit activation during training.

\begin{table}[!htb]
\centering\small\renewcommand{\arraystretch}{1.13}
\caption{\textbf{Paired-credit activation.} Eligible positions have an open gate and two valid tentative plans; nonzero positions satisfy $M_{ii}\neq M_{i,1-i}$. Training uses 384 dialogues over 24 rounds.}
\label{tab:perfamily-credit}
\begin{tabularx}{\linewidth}{@{}l*{3}{>{\raggedleft\arraybackslash}X}@{}}
\toprule
Family & Eligible & Nonzero & Nonzero rate\\
\midrule
Resource allocation & 272 & 194 & 71.3\%\\
Authorization & 180 & 118 & 65.6\%\\
Bounded quantities & 264 & 194 & 73.5\%\\
Joint scheduling & 222 & 95 & 42.8\%\\
Shared-resource scheduling & 244 & 85 & 34.8\%\\
Set-based consent & 196 & 80 & 40.8\%\\
\midrule
All families & 1378 & 766 & 55.6\%\\
\bottomrule
\end{tabularx}
\end{table}

Activation is lowest for scheduling: both tentative plans are frequently infeasible, producing a tie ($M_{ii}=M_{i,1-i}=0$) and zero paired contrast.

\section{Robustness and Human Validation}
\label{app:human-validation}

\paragraph{Generation and partner variation.}
The self-play experiment varies both participants and repeats generation with three seeds, holding the scenario set fixed. Figure~\ref{fig:robustness-results} (left) reports Verified Success, and Table~\ref{tab:selfplay} gives the corresponding Feasibility and Integrity means and standard deviations.

\subsection{Cross-Judge Agreement}
\label{app:judge-agreement}

We re-evaluate fixed model outputs with Kimi-K3 \citep{moonshot2026kimik3}, retaining the Qwen3.8-Max Integrity messages, rubric, and evidence checks. Agreement between Qwen3.8-Max and Kimi-K3 is computed on the final binary Integrity decision for each model--task evaluation.

\paragraph{Subset selection.}
A task position identifies a scenario, world, and tested role. The eligible pool is the intersection of positions with an available frozen Integrity request for Base, PPO, PaCPO, and GRPO ($G=4$): 239 positions. We stratify by the six interaction families, sort family names and then task keys lexicographically, and shuffle each family's keys using one Python random-number generator initialized with seed 20260924. Selecting the first ten keys per family without replacement yields 60 shared positions spanning 43 scenarios. Selection uses task identifiers and family labels, not Feasibility or Integrity scores, and is fixed before Kimi-K3 is queried. We reuse each method's output at these same positions without regenerating dialogues, giving 60 evaluations per method and 240 in total. Identical requests are scored once and reused, producing 203 unique Kimi-K3 requests; reported statistics give equal weight to all 240 model--task evaluations.

\begin{table}[!htb]
\centering\small\setlength{\tabcolsep}{4pt}\renewcommand{\arraystretch}{1.1}
\caption{\textbf{Qwen3.8-Max--Kimi-K3 Integrity agreement.} Both judges evaluate the same 240 model--task cases. Agreement compares final pass/fail decisions; pass rates show each judge's marginal acceptance rate.}
\label{tab:judge-agreement}
\begin{tabularx}{\linewidth}{@{}l*{4}{>{\raggedleft\arraybackslash}X}@{}}
\toprule
Judge pair & \shortstack[r]{Agreement\\(\%)} & Cohen's $\kappa$ & \shortstack[r]{Qwen3.8-Max\\pass (\%)} & \shortstack[r]{Kimi-K3\\pass (\%)}\\
\midrule
Qwen3.8-Max / Kimi-K3 & \textbf{90.00} & \textbf{0.783} & 30.00 & 40.00\\
\bottomrule
\end{tabularx}
\end{table}

Qwen3.8-Max and Kimi-K3 agree on 216 of 240 decisions, with $\kappa=0.783$ (Table~\ref{tab:judge-agreement}). Kimi-K3 accepts 24 additional cases and rejects none that Qwen3.8-Max accepts.

\paragraph{Scoring interface.}
The recorded model identifiers are \texttt{qwen3.8-max} and \texttt{kimi-k3}. Qwen3.8-Max uses strict JSON-schema output with thinking disabled. Kimi-K3 uses JSON-object output with high reasoning effort and local validation against the same frozen schema. Field normalization renames \texttt{explanation} to \texttt{reason}, removes redundant speaker fields from evidence, and fills omitted empty fields for negative flags. Format-repair prompts instruct the judge to retain substantive judgments and cited evidence. All 240 evaluations per judge yield valid final decisions; original responses and repairs are retained.

\subsection{Human Review}
\label{app:human-review}

\paragraph{Protocol.}
Two expert annotators review 48 dialogues from Base and \method{},
covering 12 scenarios and both of their worlds, which form 24 matched
Base--\method{} pairs. For each dialogue, annotators see the public
context, both role cards, the full conversation, and both records, with
model identities hidden. They judge whether the final plan satisfies the
stated conditions (Feasibility) and whether the participants' utterances
support it (Integrity). For each matched pair, presented in random order
\citep{zheng2023judging}, they also judge which model provides more
adequate help to the partner, citing turn-level evidence.

\paragraph{Results.}
Human Feasibility judgments agree with the executable checks on 100\% of
dialogues, and human Integrity judgments agree with Qwen3.8-Max on
82.22\%. In the helpfulness comparison, annotators prefer \method{} in 8
pairs and Base in 5, rate the two as similar in 3, and judge both
inadequate in 8. \method{} thus receives more favorable judgments, while
one third of the pairs reveal cases in which neither model helps
adequately.

\section{Limitations and Future Work}
\label{app:limitations}
\label{sec:limitations}

\paragraph{Scope of outcome verification.}
HICO-Bench covers six coordination families whose arrangements can be represented as structured records and checked against explicit conditions. This supports reproducible feasibility tests, but does not cover the full range of social interaction, such as emotional support or relationships without a well-defined final outcome. Extending the benchmark requires identifying verifiable outcomes without reducing these interactions to inappropriate formal objectives. Feasibility success is feasibility under the specified conditions, not a general judgment of social appropriateness.
\end{document}